%% file: acl_latex.tex
\pdfoutput=1

\documentclass[11pt]{article}
\usepackage{authblk}
\usepackage[final]{acl}

\usepackage{graphicx}
\usepackage{amsmath,amssymb} 
\usepackage{makecell}
\usepackage{color}
\usepackage{xcolor,colortbl}
\usepackage{enumitem}
\usepackage{booktabs}
\usepackage{float}
\usepackage{tabulary,multirow,overpic,xcolor}
\usepackage{verbatim}
\usepackage{enumitem}
\usepackage{array}
\usepackage{arydshln}
\usepackage{adjustbox}
\usepackage{siunitx}
\usepackage{algorithm}
\usepackage{algpseudocode}
\usepackage{times}
\usepackage{latexsym}
\usepackage{mathtools}
\usepackage{wrapfig}
\usepackage{tabularx}
\usepackage{arydshln, xcolor}
\usepackage{bbm}
\usepackage{caption}
\usepackage{subcaption}

\usepackage[T1]{fontenc}

\usepackage[utf8]{inputenc}

\usepackage{microtype}

\title{Selective Token Generation for Few-shot Natural Language Generation}

\author[1]{Daejin Jo}
\author[1]{Taehwan Kwon}
\author[2]{Eun-Sol Kim}
\author[1]{Sungwoong Kim}
\affil[1]{Kakao Brain}
\affil[2]{Hanyang University}
\affil[1]{\texttt{\{daejin.jo, taehwan.kwon, swkim\}@kakaobrain.com}}
\affil[2]{\texttt{eunsolkim@hanyang.ac.kr}}

\begin{document}
\maketitle
\begin{abstract}
Natural language modeling with limited training data is a challenging problem, and many algorithms make use of large-scale pretrained language models (PLMs) for this due to its great generalization ability. Among them, additive learning that incorporates a task-specific adapter on top of the fixed large-scale PLM has been popularly used in the few-shot setting. However, this added adapter is still easy to disregard the knowledge of the PLM especially for few-shot natural language generation (NLG) since an entire sequence is usually generated by only the newly trained adapter. Therefore, in this work, we develop a novel additive learning algorithm based on reinforcement learning (RL) that selectively outputs language tokens between the task-general PLM and the task-specific adapter during both training and inference. This output token selection over the two generators allows the adapter to take into account solely the task-relevant parts in sequence generation, and therefore makes it more robust to overfitting as well as more stable in RL training. In addition, to obtain the complementary adapter from the PLM for each few-shot task, we exploit a separate selecting module that is also simultaneously trained using RL. Experimental results on various few-shot NLG tasks including question answering, data-to-text generation and text summarization demonstrate that the proposed selective token generation significantly outperforms the previous additive learning algorithms based on the PLMs.

\end{abstract}

\section{Introduction}


Recently, pretrained language models (PLMs) have shown great generalization ability when combined with large-scale data and big transformer-based models \citep{devlin-etal-2019-bert, radford2019language, Mike2020bart, Brown2020gpt3, KalyanAMMUS, petroni-etal-2019-language, wang2020language}. 
Therefore, transfer learning from PLMs has been popularly used for few-shot natural language generation (NLG) tasks with promising results. In specific, the use of PLM for few-shot NLG can be categorized into three approaches: 1) prompt-based, 2) finetuning, and 3) additive learning. Prompt-based approaches encode a task description and task-specific examples as a natural language prompt for few-shot text generation \citep{radford2019language, Brown2020gpt3, Zheng2021, Schick2020, Lisa2021}. 
These approaches can take full advantage of the universal natural language understanding and generation capabilities of large-scale PLMs without further training of the main model, however, they have some limitations in dealing with a large domain shift from the pretraining corpus data, tuning suitable task-specific prompts, and covering an increased size of conditioning examples. 
On the other hand, finetuning of the PLM is able to explicitly impart task-specific knowledge to the model and hence lift the above limitations \citep{Ziegler2019, AUGNLG, chen-etal-2020-shot}. However, these finetuned models are prone to overfitting when only a small amount of training data is available. In order to alleviate such an overfitting problem, additive learning has been extensively exploited by incorporating task-specific adapters into the PLM \citep{Asa2019, houlsby2019parameter, zeldes2020technical}. 

\input{./exp_data/qa_example_intro}

In general, task-specialized adapters for few-shot NLG are trained by maximum likelihood estimation (MLE) or reinforcement learning (RL). 
While MLE is efficient in learning, it suffers from the exposure bias problem due to the difference in the training and inference mechanisms \citep{Tianxing2019}, and this problem can be severe with limited training data.
One solution is RL, capable of resolving this exposure bias problem by sequential output sampling during training \citep{ranzato2015sequence, keneshloo2019deep, shi2021neural}.
However, the exponentially large space of output sequences restricts the use of RL since it leads to high variance and unstable training which is more serious in the few-shot setting.

More importantly, the existing additive learning generally produces the whole output sequence by its own task-specific adapter, which leads to a fundamental limitation in maintaining the knowledge of the PLM and the strong generation ability. An example of this limitation from our empirical observation on the task of question and answering is shown in Table \ref{tab:qa_example_intro}. In this case, a passage that contains two definitions (super-scripted and bolded) about \textit{conflict} is given with a query that asks about the psychological meaning of \textit{conflict}.
Without the knowledge of \textit{who Colman\footnote{A psychologist, \url{https://en.wikipedia.org/wiki/Peter_T._Coleman_(academic)}} is}, it can be hard to answer since the word \textit{psychology} in the query does not appear in the passage.
Here, the PLM repeats the given query as its generated answer due to the lack of domain adaptation while the added adapter incorrectly outputs not the psychological meaning but the general meaning of \textit{conflict}. This is because most queries in this few-shot training data ask a general meaning of a concept, and therefore the adapter is overfitted to this pattern (more examples are described in Section \ref{section:result}). Note that the PLM generates the correct answer if the proper conditioning text (\textit{the meaning of conflict is}) is provided.

Motivated by these observations, in this work, we propose a novel RL-based selective token generation (STG) between the task-general PLM and the task-specific adapter. The selection of this output token generator enables to explicitly maintain a general prior knowledge from the frozen PLM and the adapter to focus only on the task-relevant parts in sequence generation. Note that the proposed algorithm is different from previous selective generation algorithms such as \textit{copy mechanism} \citep{gu-etal-2016-incorporating} in that STG selects a generator rather than existing tokens in a given passage.
In few-shot learning, the proposed partial token generation makes the task-specific adapter more resilient to overfitting and furthermore reduces the overall output space which leads to stable RL training. Here, in order to make the two token generators (policies) complement each other as well as to realize the robust output selection at the token level on the fly, we exploit a separate token-level policy selector. 
Note that both the policy selector and the task-specific adapter are simultaneously learned by the RL algorithm.
Experimental results on various few-shot NLG tasks show that the proposed selective token generation outperforms the previous PLM-based additive learning algorithms with the comprehensive (non-selective) token generation.



Our main contributions can be summarized as follows.
\begin{itemize}
    \setlength\itemsep{0em}
    \item A novel selective token generation between the PLM and the task-specific adapter is proposed for few-shot NLG.
    \item RL is applied to train both the policy selector and the task-specific adapter that is complementary to the PLM in text generation.
    \item Extensive empirical validation on few-shot NLG tasks demonstrates that the proposed selective token generation performs better in comparison to the previous PLM-based additive learning algorithms.
\end{itemize}

\section{Background}
\subsection{Natural Language Generation}
The goal of NLG is to generate a text sequence ${\bf y} = [y_0, ..., y_T]$ for a given task, where $y_t$ is the $t$th output token from a vocabulary $\mathcal V$, and $T$ is the output sequence length. For this generation, we aims to model the distribution of $\bf y$ that is autoregressively factorized as $p_{\theta}({\bf y}) = \prod_{t=0}^{T} p_{\theta}(y_t | {\bf y}_{<t})$, where $\theta$ denotes the model parameters and ${\bf y}_{<t} = [y_0, ..., y_{t-1}]$. Here, the conditional distribution to sample a token for each step, $p_{\theta}(y_t | {\bf y}_{<t})$, is defined by the softmax function on the output logits $f_{\theta}(y_t | {\bf y}_{<t})$. Note that in general, the language generation is conditioned on input context according to a given task. Here, we encode the conditioning context by the same sequential model for generating an output sequence, and for simplicity we omit it. 

\begin{figure*}[t]
    \centering
    \includegraphics[width=0.8\textwidth]{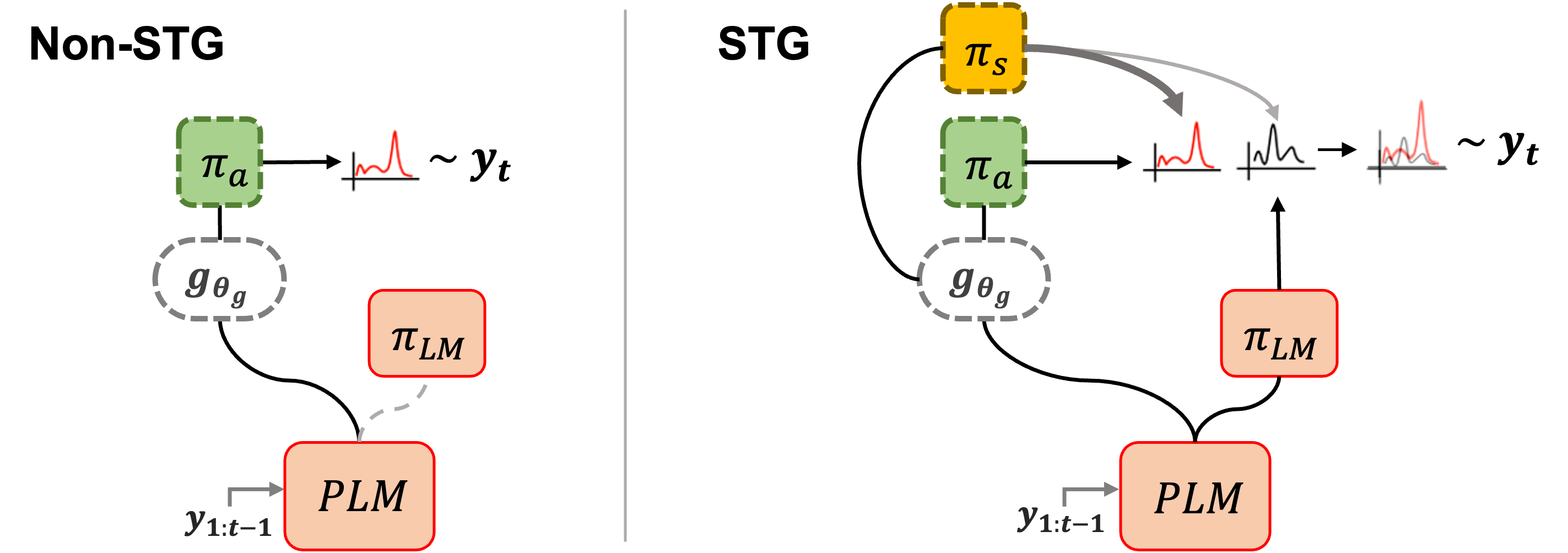}
    \caption{Text generation processes of Non-STG and STG are described. In the Non-STG, every token is sampled from the task-specific policy $\pi_a$ (\textbf{Left}). On the other hand, in the proposed STG, each token is selectively sampled from either the PLM policy $\pi_{LM}$ or the test-specific policy $\pi_a$ where the selection is performed by the selection policy $\pi_s$ (\textbf{Right}). Symbols with dashed line represent learnable models.}
    \label{fig:main}
\end{figure*}

\subsection{Additive Learning for Few-shot Generation} \label{sec:2.2}
To effectively leverage the general linguistic knowledge, $\theta$ is first initialized by the PLM parameters, $\theta_{LM}$, for NLG. Given $N$ task-specific training instances, ${\mathcal D}=\{{\bf y}^{n*}\}_{n=1}^{N}$, where ${\bf y}^{n*}$ is the $n$th ground-truth output sequence, directly finetuning $\theta_{LM}$ using ${\mathcal D}$ can incur the severe overfitting problem when $N$ is small in the few-shot scenario. 
Therefore, we add the task-specific adapter, $g_{\theta_{a}}$ parameterized by $\theta_{a}$, on top of the PLM, and optimize only $\theta_{a}$ \citep{zeldes2020technical, Asa2019}. 
In specific, we reformulate $f(\cdot | {\bf y}_{<t}; \theta) = W^T h({\bf y}_{<t}; \theta_{h})$
where $W \in {R}^{H \times |\mathcal V|}$ and $h \in {R}^{H}$ denote the weight matrix and the penultimate representations, respectively, and $\theta = \{W, \theta_{h}\}$. 
Then, we define the task-specific conditional distribution as follows:
\begin{multline} \label{eq:1}
p (y_t | {\bf y}_{<t}; \theta_{LM}, \theta_{a}) = 
\sigma
\bigg ( {W_{LM}}^T h_{LM} ({\bf y}_{<t}) \\
+ {W_{a}}^T g \big ( h_{LM} ({\bf y}_{<t}); \theta_{g} \big ) \bigg ),
\end{multline}
where $h_{LM} ({\bf y}_{<t}) = h({\bf y}_{<t}; \theta_{h, LM})$, $\theta_{a} = \{W_a, \theta_{g}\}$ and $\sigma$ is the softmax function. Here, the summation of the PLM logits and the adapter logits is motivated by auxiliary training\footnote{Although the auxiliary training is particularly designed for maximizing the likelihood of the target task output, it also can take an advantage for RL since the adapter logits are nearly zero before training is advanced. Namely, it lets the task-specific conditional distribution start learning from the distribution of PLM, not a uniform distribution.} \citep{zeldes2020technical}. 
It is noted that in our additive learning $\theta_{a}$ is updated while $\theta_{LM}$ is kept frozen. Hence, in the following we omit $\theta_{LM}$ such that $p_{\theta_{a}}(y_t | {\bf y}_{<t}) = p (y_t | {\bf y}_{<t}; \theta_{LM}, \theta_{a})$ for simplicity. 


\subsection{Reinforcement Learning (RL)} \label{sec:2.4}
As an alternative to MLE, RL is able to overcome the exposure bias problem of MLE by sequence-level sampling from the model distribution during training \citep{ranzato2015sequence} and allows to leverage the target-specific sequence-level objectives such as BLEU \citep{Lijun2018, Guo2021}.
In order to use RL for our additive learning, we reformulate our text generation as an RL problem: at each time step $t$, the agent takes the current state ${\bf s}_t = {\bf y}_{<t}$ as an input and performs an action $a_t$ that outputs a token $y_t$  by a policy $\pi_{\theta} (a_t|{\bf s}_t)$ corresponding to $p_{\theta}(y_t | {\bf y}_{<t})$. 
Then, the agent receives a reward $r_t = r({\bf s}_t, a_t)$ and deterministically transitions to the next state ${\bf s}_{t+1}$. 
Here, note that the token-level intermediate reward $r_{t} = 0, \forall t < T$ when we use the delayed reward associated with the sequence-level evaluation metric between the two full sequences, $\bf y$ and ${\bf y}^*$. 
Let $\tau= \{({\bf s}_t, a_t, r_t)\}_{t=0}^{T}$ be the trajectory generated by $\pi_{\theta}$. 
The RL objective for the optimal agent is to maximize the expected sum of future discounted rewards ${\mathbb E}_{\tau \sim \pi_{\theta}} [ \sum_{t=0}^{T} \gamma^t r_t ]$,
where $\gamma \in [0,1]$ is the discount factor. We employ an actor-critic algorithm \citep{bahdanau2017actor} which requires the additional critic network to estimate the value of a state, $V^{\pi}({\bf s}_t) = {E}_{\pi}[\sum_{t'=t}^{T} \gamma^{t'-t} r_{t'}|{\bf s}_t] = \sum_{a_t} \pi(a_t|{\bf s}_t)Q^{\pi}({\bf s}_t, a_t)$ where the state-action value function $Q^{\pi}({\bf s}_t, a_t) = {E}_{\pi}[\sum_{t'=t}^{T} \gamma^{t'-t} r_{t'}|{\bf s}_t, a_t] = r_t + V^{\pi}({\bf s}_{t+1})$.
We use the policy gradient loss to learn the policy parameters $\theta$:
${\mathcal L} = - \sum_{t=0}^{T} A^{\pi_{\theta}}({\bf s}_t, a_t) \log \pi_{\theta}(a_t | {\bf s}_{t}),$
where $A^{\pi_{\theta}}({\bf s}_t, a_t) = Q^{\pi_{\theta}}({\bf s}_t, a_t) - V^{\pi_{\theta}}({\bf s}_t)$ is the advantage function.

\section{Selective Token Generation}
Instead of generating all tokens in an output sequence from the single task-specific policy, $\pi_a = \pi_{\theta_a}(a_t|{\bf s}_t)$, at each time step $t$, we sample an output token $y_t$ selectively from either the PLM policy $\pi_{LM} = \pi_{\theta_{LM}}(a_t|{\bf s}_t)$ or the task-specific policy $\pi_{a}$:
\begin{multline} 
y_t = a_t \sim \big ( \mathbbm{1}_t[\text{$\pi_{LM}$ is selected}] \pi_{LM}(a_t|{\bf s}_t) \\
+ (1-\mathbbm{1}_t[\text{$\pi_{LM}$ is selected}]) \pi_{a}(a_t|{\bf s}_t) \big ),
\end{multline}
where $\mathbbm{1}_t[\cdot]$ is the indicator function (at $t$) that equals 1 if it is true and 0 otherwise.
This output token selection allows to explicitly utilize a general linguistic knowledge from the PLM without catastrophic forgetting in few-shot learning. Also, the task-specific policy can focus on generating only the task-relevant parts, which enables more effective few-shot training with a reduced search space.

Now we need to determine how to select the proper policy at each step on the fly as well as to make the task-specific policy complementary to the PLM policy. For this, we exploit a separate token-level policy selector. The proposed policy selector $\pi_{s}(i_t|{\bf s}_t; \theta_s)$ with the parameters $\theta_s$, where $i_t \in \{0, 1\}$, is an another policy that stochastically decides a policy to generate $a_t$ for ${\bf s}_t$. Namely, a token sample $y_t$ is generated by the following process:

\begin{eqnarray}
i_t &\sim& \pi_{s}(i_t|{\bf s}_t), \\
y_t &=& 
\begin{cases}
a_t\sim\pi_{LM}(a_t|{\bf s}_t) & \text{if} \ i_t = 0,   \\
a_t\sim\pi_{a}(a_t|{\bf s}_t) & \text{if} \ i_t = 1.
\end{cases}
\end{eqnarray}
This process can be considered as a token generation from a hierarchical policy $\pi_h(a_t|{\bf s}_t; \theta_s, \theta_{LM}, \theta_a)$ where the policy selector represents the upper-level prior for the preference of the low-level policy. Therefore, the value function of this hierarchical policy can be formulated as
\begin{eqnarray*}
&V^{\pi_h}& = {\mathbb E}_{\pi_h}[\sum_{t'=t}^{T} \gamma^{t'-t} r_{t'}|{\bf s}_t] \\
&=& \pi_{s}(0_t|{\bf s}_t) \sum_{a_t} \pi_{LM}(a_t|{\bf s}_t)Q^{\pi_{h}}({\bf s}_t, a_t) \nonumber\\
&+& \nonumber \pi_{s}(1_t|{\bf s}_t) \sum_{a_t} \pi_{a}(a_t|{\bf s}_t)Q^{\pi_{h}}({\bf s}_t, a_t), \nonumber\\
\end{eqnarray*}
and $A^{\pi_h}({\bf s}_t, a_t) = Q^{\pi_{h}}({\bf s}_t, a_t) - V^{\pi_h}({\bf s}_t)$. 
We denote $i_t=0$ and $i_t=1$ as $0_t$ and $1_t$ respectively.
Here, it is noted that a single critic network is used for the hierarchical policy since $i_t$ does not affect ${\bf s}_t$. Given a sample trajectory $\{({\bf s}_t, i_t, a_t, r_t)\}_{t=0}^T$, the loss for optimizing $\theta_s$ and $\theta_a$ is
\begin{eqnarray} \label{eq:9}
 {\mathcal L} = - \sum_{t=0}^{T} A^{\pi_h}({\bf s}_t, a_t) 
\bigg ( \mathbbm{1}[0_t] \mathcal L_{LM}
+ \mathbbm{1}[1_t] \mathcal L_{a}
\bigg ),
\end{eqnarray}
where
\begin{align*}
{\mathcal L_{LM}} =& \log sg[\pi_{LM}(a_t | {\bf s}_{t})] + \log \pi_{s}(i_t | {\bf s}_{t}),  \nonumber \\
{\mathcal L_{a}} =& \log \pi_{a}(a_t | {\bf s}_{t}) + \log \pi_{s}(i_t | {\bf s}_{t})
\end{align*}
and \textit{sg} stands for the stop-gradient operator.
Similar to $\pi_a$, $\pi_s$ makes use of the PLM representations and the task-specific adapter such that
\begin{eqnarray} \label{eq:6}
\pi_s(i_t | {\bf s}_t; \theta_s) = \sigma
\bigg ( m \Big ( g \big ( h_{LM} ({\bf s}_{t})\big ); \theta_s \Big ) \bigg ),
\end{eqnarray}
where $\sigma$ is the softmax function and $m$ is the selector module. Figure \ref{fig:main} depicts the overall text generation process by the proposed selective token generation (STG) in comparison to the previous non-selective token generation (Non-STG). Here, note that since all policies in STG share the same PLM representations, the increased computational cost by STG over Non-STG is negligible.

The use of the separated policy selector that is simultaneously trained with the task-specific policy allows the task-specific policy to be complementary to the PLM policy. 
Especially, this cooperative ensemble learning can be realized by our RL algorithm that performs sequential sampling from the model during training. 

\begin{figure}[t]
\begin{minipage}{.45\textwidth}\centering
    \includegraphics[width=1\textwidth]{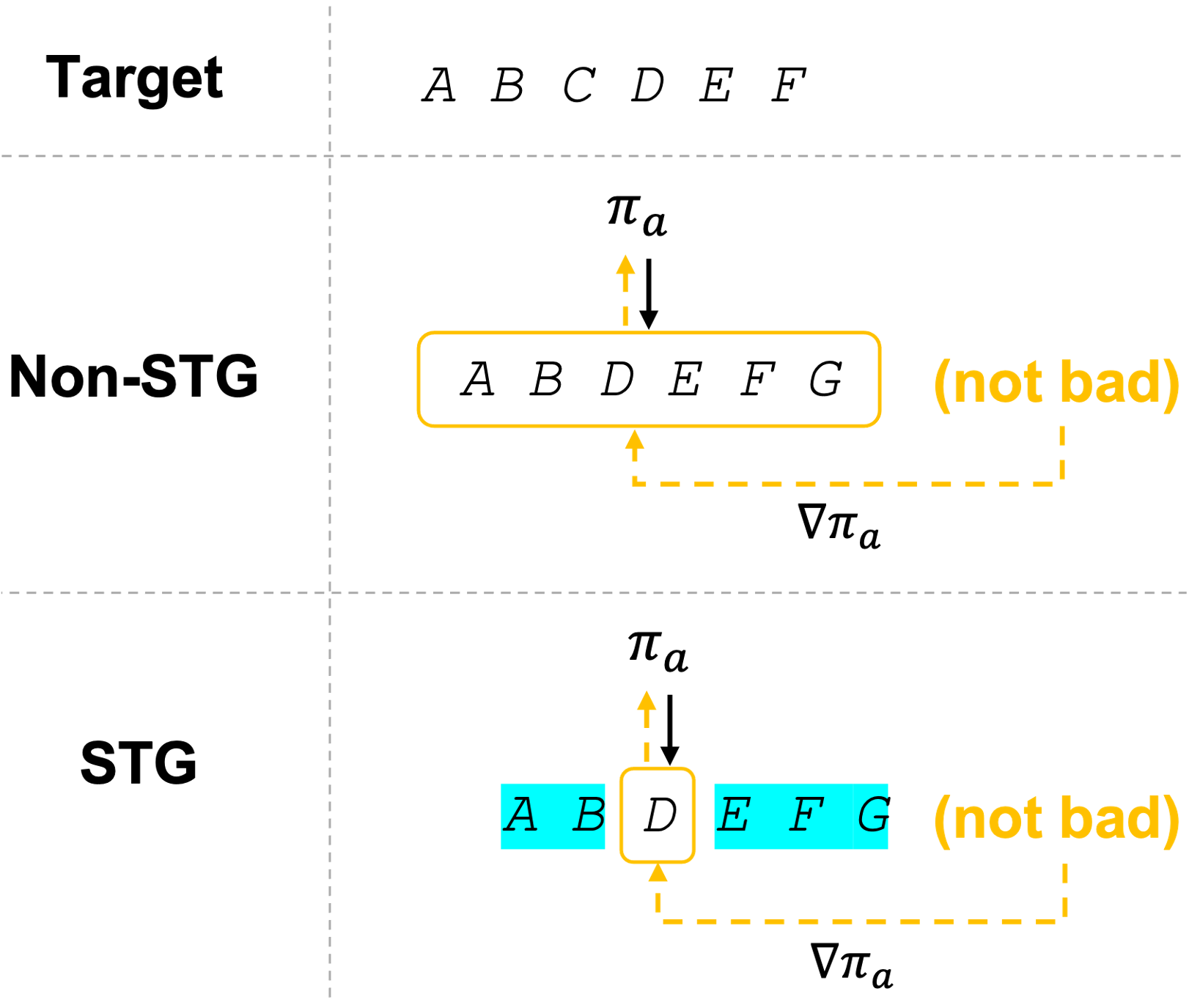}
    \caption{\label{fig:illust_new}A simple schematic illustration of Non-STG and STG. \textbf{Non-STG(RL):} the whole sequence of target is generated from the task-specific policy $\pi_a$ so the right sub-sequence {\fontfamily{qcr}\selectfont AB} is also penalized from the delayed feedback. \textbf{STG:} the third token is sampled from $\pi_a$ and the model lets the other tokens (highlighted with \textcolor{cyan}{cyan}) generated from the PLM's policy $\pi_{LM}$ which generates a next letter of the previous alphabet input. Here, $\pi_a$ will be penalized at the third token.}
\end{minipage}
\end{figure}

The advantages of STG are as follows:
\textbf{(1)} STG makes use of the PLM not at the feature level but the output distribution level in text generation. In our few-shot learning this is beneficial in explicitly retaining strong linguistic and world knowledge from the PLM.
\textbf{(2)} STG resolves the exponentially large search space $|\mathcal V|^{T}$ since the frozen PLM chooses a token when it is selected, and therefore the search space of the generator is approximately decreased from $|\mathcal V|^{T}$ to $|\mathcal V|^{T-\overline{T}_{PLM}}$ where $\overline{T}_{PLM}$ is the average length of sequences generated by PLM. 
\textbf{(3)} STG is efficient in credit assignment. The loss function of STG (Equation \ref{eq:9}) intuitively shows that the gradient to the task-specific policy $\pi_a$ associated with producing $a_t$ will depend on the selector's action (i.e. $i_t=1$). Hence, unlike Non-STG, $\pi_a$ of STG knows which token is used as a task-specific token and contributed to the reward (see Figure \ref{fig:illust_new} for an illustration).

It is noted that although the STG also can be trained by MLE, it can be easily collapsed to select only a task-specific policy irrespective of a given content. We analyze the MLE version of STG in Appendix \ref{sec:study}.

\section{Experiments}
We evaluate our method against additive learning baselines on Data-to-Text, Question Answering and Text Summarization tasks which are widely used in few-shot NLG. 
\subsection{Baseline}
\noindent\textbf{PLM}. 
In our experiments, we assume that the PLM works to some extent for a given task. 
However, the naive PLM usually does not satisfy it for a new task unseen during training. 
Hence, we finetuned GPT-2\footnote{We make use of GPT-2 with 345M parameters as the initial checkpoint. We follow the training details in the previous works \citep{peng2020few, khandelwal2019sample} for each task.} \citep{radford2019language} with MLE for few epochs and used it as the PLM. 
Fine-tuning the PLM with MLE is most commonly used for task adaptation and thus it can also be a strong baseline.
This fine-tuning phase accelerates the learning of the adapter. This is particularly when the adaptation requires to cover the large domain shift. Severe performance degradation was observed for all the tasks when we skipped this fine-tuning.

\noindent\textbf{Non-STG}. 
This method stands for Non-Selective Token Generation which uses the above the PLM as an encoder (frozen) and the adapter (additional layer to be trained).
We use two objectives, MLE and RL, for additive learning. 
These will be denoted as Non-STG-MLE and Non-STG-RL, respectively. 

\input{./exp_data/tod}

\noindent\textbf{STG-Naive Ensemble}. 
We believe that the proposed generation encourages the task-specific policy ($\pi_a$) to complement the PLM's policy ($\pi_{LM}$) with a proper selection of the selector through the joint training. 
To investigate this, we evaluate against two different naive ensembles of the policies, $\pi_{a}$ trained from Non-STG and $\pi_{LM}$ of the PLM. 
These ensemble schemes are as follows: 
\begin{itemize}
\setlength\itemsep{0em}
\item NE($max$): $\pi_{max} = \sigma(Max(\pi_{a}, \pi_{LM}))$
\item NE($mix$): $\pi_{mix} = (\pi_{a} +\pi_{LM})/2$
\end{itemize}
We also evaluate another naive ensemble strategy NE($random$) that randomly selects a token policy at each step between $\pi_{a}$ and $\pi_{LM}$, however it shows lower performances than the others.


\subsection{Implementation} \label{Implementation}
\noindent\textbf{Adapter}.
The task-specific adapter $g$ in Section~\ref{sec:2.2} is implemented by a LSTM to encode the dynamics of the representation vector $h_{LM}$. We found that the use of MLP was not good in the sense of performance.


\noindent\textbf{Selector}. 
We use a 2-layer MLP with ReLU activation for $m$ of Equation \ref{eq:6}.


\input{./exp_data/qa_1}

\noindent\textbf{Reinforcement Learning}. 
We employ Actor-Critic method \citep{konda2000actor, fedus2018maskgan} for RL.
The agents (i.e. selector and generator) receive a reward after generating a sentence. 
Here, we use different reward functions according to tasks. 
We use delexicalised BLEU for Data-to-Text following \cite{peng2020few}, Averaged score of BLEU and ROUGE-L for Question Answering and ROUGE-L for Text Summarization following \cite{paulus2017deep} as the reward function. 

\noindent\textbf{Token Sampling}. 
During the training, $i_t \in \{0, 1\} \sim \pi_s$ is first sampled, and then we use either $\pi_{LM}$ of the PLM for $i_t=0$ or the task-specific policy $\pi_a$ for $i_t=1$ to sample the $t$th token.
During the evaluation, any decoding strategy, such as a beam search, can be used with the mixture of policies $\pi_{h}(\cdot) = \pi_s(0_t)\pi_{LM}(\cdot) + \pi_s(1_t)\pi_a(\cdot)$.
We use the beam search decoding with a sample size of $k=3$ for Text Summarization and $top_p=0.9$ decoding for both Data-to-Text ($k=10$) and Question Answering ($k=3$). 

\subsection{Data-to-Text}
Data-to-Text is a task that transforms structured data such as graphs or tables into natural language. 
Recent works \citep{mager2020gpt, peng2020few, kale2020text} show that the PLM can be adapted successfully to this task by taking a serialized form of data as an input without a carefully designed model to encode the structured data. 
Here, we perform experiments on FewShotWOZ \citep{peng2020few} dataset. The evaluation is conducted on the topics which are available\footnote{ \url{https://github.com/pengbaolin/SC-GPT}}. 
Only 50 instances for each topic are available for training and 129, 78, 1379, and 680 testing instances for Restaurant, Hotel, Laptop, and TV, respectively. 
The models are evaluated by measuring fluency and informativeness using BLEU score and ERR (slot ERror Rate), respectively. Table \ref{tab:tod} shows the obtained results.

\subsection{Long Answer Question Answering}
We consider Long Answer Question Answering (QA) task on MS-MARCO \citep{nguyen2016ms} dataset. In this task, a passage and a query are given, and the model generates an answer with respect to the query by referring to the passage. Here, we randomly sample various sizes of (50, 100, 500, 1,000 $\approx 1\%$, and 2,000) subset data from the train dataset. We also sample a validation and a test set, which contains 500 and 12,000 instances, respectively, from the dev dataset.
We repeat this test three times with different random seeds and thus perform experiments on total nine subsets.
The models are evaluated by measuring BLEU and ROUGE-L (denoted as R-L). 
We report \textit{averaged performances} over the three runs and \textit{averaged performance gain} against the PLM in Table \ref{tab:qa} and Figure \ref{fig:QA_gain}, respectively.

\subsection{Text Summarization}
We consider the problem of abstractive summarization for long text generation.
Here, we randomly sample various sizes of (50, 100, 300, 1,500, and 3,000 $\approx 1\%$) subset data from CNN/Daily Mail \citep{see2017get}.
We repeat this test three times for each size of few-shot as in above QA task. 
ROUGE \citep{lin-2004-rouge} is commonly used to evaluate n-grams recall of the summaries with gold references.
The models are evaluated by measuring ROUGE-1, ROUGE-2, and ROUGE-L (denoted as R1, R2, and R-L, respectively). 
We report \textit{averaged performances} over the three runs and \textit{averaged performance gain} against the PLM in Table \ref{tab:summ1} and Figure \ref{fig:summ_gain}, respectively.

\input{./exp_data/summ_1}

\subsection{Result} \label{section:result}
In most cases, additive learning improves the performances over the PLM. 
However, they do not always guarantee a performance improvement. 
For example, the ERR score of the PLM on \textit{Laptop} shows a better result except for STG and NE($mix$)-RL (see Table~\ref{tab:tod}) and the Non-STGs trained on $1,000 \approx 1\%$ few-shot subset of MS-MARCO do not outperform the PLM (see Table \ref{tab:qa}). 

\noindent\textbf{Data-to-Text}. As shown in Table~\ref{tab:tod}, we can observe that the Non-STGs do not outperform the PLM even though it has more neural units and takes more training time.
The models trained on the RL objective show better performances for the ERR (lower is better). 
Interestingly, NE($mix$) methods show strong improvements for the BLEU which measures the fluency of sentence but obvious degeneration for the ERR which measures the rate of missing information from the given data. 
These results suggest that the PLM is much more capable of \textit{task-general} knowledge than the \textit{task-specific} generator (i.e. $\pi_a$) trained on few-shot dataset, which ensures our motivation of jointly training the policy selector and the task-specific generator is valid.
Note that while other methods show some trade-off between BLEU and ERR, only STG shows improvements on both metrics for all topics in the dataset.

\noindent\textbf{Question Answering}. As shown in Table \ref{tab:qa},  STG shows significantly better performances than the other methods.
Notably, NE($mix$) show good performances as much as STG especially where the training data size $\ge 1,000$. 
It obviously suggests that the PLM can be a complementary model to the additional model. Therefore, in this context, it can be lost of the prior knowledge of the PLM even if the additional model has been built over the feature space of the PLM.
In addition, we can expect that STG would be more beneficial on the small number of samples for this kind of tasks which depend on the PLM's ability like common sense knowledge.
As shown in Figure \ref{fig:QA_gain}, STG shows strong improvements compared to Non-STG-RL especially where the training data size $\le 500$.


\noindent\textbf{Summarization}. As shown in Table~\ref{tab:summ1}, STG shows significantly larger gains than Non-STGs, and their naive ensembles with the PLM in every score metric and training data size.
Similar to QA, STG shows improvements compared to Non-STGs especially where the training data size $\le 300$ as shown in Figure \ref{fig:summ_gain}.
However in contrast to the QA task, the improvement may seem limited for all models including STG. 
We think that the adapters used in this study may not be suitable for this particular task which requires to understand the long context and compress it into a summary.
It may need the use of lower-level features or more parameters to adapt to such tasks.
We discuss this limitation in Section \ref{sec:limit}.

\noindent\textbf{Overfitting in Non-STGs}. In the example as shown in Table \ref{tab:qa_example_intro} the answer of STG, which is close to the ground truth, is generated by the PLM policy $\pi_{LM}$ after some sequence of tokens (\textit{conflict is}) that are sampled from the task-specific policy $\pi_a$. 
The Non-STGs generate general meaning which is not intended.  
We can find such examples for the other tasks in Appendix \ref{sec:gen_sents}:
In Data-To-Text, as shown in the last example of Table \ref{tab:tod_exam}, Non-STG generates \textit{nicam stereo} which is not appeared in the given data. This is due to that \textit{nicam stereo} was appeared 7 times (7/50, $14\%$) in training data.
In Summarization, as shown in the first example of Table \ref{tab:summ_exam2}, Non-STGs only consider the forepart of the given article. Since the most of the major information is appeared in the forepart in News data, Non-STGs can be easily overfitted to generate the text according to such a pattern.
Hence, we claim that Non-STG is easily exposed to learning patterns of typical answering, but STG resolves this issue since it can be fully accessible to the knowledge of the PLM.

\section{Related Work}
Recently, prompt-based in-context learning with an extremely large PLM shows impressive few-shot generation performances \citep{radford2019language, Brown2020gpt3}. \citet{Schick2020} propose manually designed natural language prompts for improved few-shot text summarization and headline generation. \citet{Elsahar2018} conduct zero-shot learning for question generation from knowledge graphs, however they require a large amount of in-domain training data for their transfer learning. \citet{chen-etal-2020-shot} directly finetune the pretrained GPT-2 with a small amount of serialized attribute-value pairs for table-to-text generation. \citet{TableGPT} further apply multiple tasks to effectively leverage the structured information of tables. In contrast to these approaches, our proposed method utilizes RL-based additive learning for few-shot text generation. 

Applying RL for text generation has been widely used to mitigate the exposure bias
problem of MLE as well as to directly optimize task-relevant evaluation metrics. \citet{ranzato2015sequence} use the REINFORCE algorithm for text summarization and machine translation while \citet{bahdanau2017actor} use the actor-critic algorithm for machine translation. 
However, they require pretraining using MLE. 
\citet{Ding2017} propose softmax policy gradient to remove the MLE-based pretraining. 
However, it requires various techniques for effective training. 
\citet{Tan2018} propose an entropy-regularized policy optimization that subsumes many of the previous training algorithms. Our proposed method is different from these methods in that we apply RL for more difficult few-shot generative modeling.

The use of RL training in PLM has been explored in many works.
\citet{Dathathri2020Plug} propose a controllable text generation which uses discriminators to guide generation of the PLM. 
This approach assumes that constant \textit{classes} like topics or preferences are available.
\citet{lazaridou-etal-2020-multi} use a PLM as a caption generator for given image. In their referential game, the generator is rewarded by a kind of discriminator that responses a signal to the generator whether the corresponding caption is correct or not.

Various methods take into account the RL tasks with large action spaces like NLG. \citet{dulac2015deep} consider only actions in a cluster around the latent state of action obtained from a given state. \citet{chandak2019learning} define the action embedding as a distribution with semantic of action and use a deterministic policy to take an action. \citet{even2003action, zahavy2018learn} devise a method of incorporating the process of directly removing unnecessary actions according to the state in the RL problem. 
Unlike these approaches, we use the hierarchical policy that reduces the sequential action space.

\section{Limitations \& Future work}\label{sec:limit}
\noindent\textbf{Adapter}. 
In this study, we aim to propose a new generation framework for few-shot natural language generation tasks.
In particular, a relatively naive neural adapter which utilizes only the top layer of the PLM is used in this paper, and thus it may lead to limited improvements as shown in the experimental results on the summarization task.
Fortunately, there are several neural architectures \citep{houlsby2019parameter, li-liang-2021-prefix, flamingo} for efficient task adaptation, and we believe that such adapters also make STG more efficient for covering a large domain shift and scaling.
The study on the architectures of the adapters will be conducted in future works.

\noindent\textbf{Efficient exploration}. 
The fundamental limitation in STG is a high dependency on PLM; 
When STG has a sufficient powerful PLM, the selector does not select the additional adapter and it is thus nothing more than the PLM. 
We can find such phenomenon in some examples in Table \ref{tab:qa_exam} and \ref{tab:qa_exam2} in Appendix. 
On the other hand, when STG has an extremely poor PLM, the selector selects the adapter always and it is thus equivalent to Non-STG. 
Therefore, in the perspective of exploration of RL the STG needs balanced selections between the PLM and the adapter. 
Furthermore, the use of RL objective requires more training time than the methods which use MLE objective such as Prefix-Tuning \cite{li-liang-2021-prefix} due to the auto-regressive sequence sampling during training.
Therefore, an analysis on efficient exploration of STG is important for future works.

\section{Conclusion}
In this work, we propose to exploit a selective token generation between the pretrained language model and the task-specific adapter with RL-based additive learning for the tasks of few-shot natural language generation. In particular, we devise a trainable policy selector at the token level and jointly learn it with the task-specific policy. The proposed policy selector and RL algorithm make the two policies complementary to each other and lead to robust few-shot generative modeling. Experimental results on various tasks of few-shot text generation show that the proposed selective token generation along with RL-based additive learning consistently and significantly improves the performances with less overfitting. 

\bibliography{anthology,custom}
\bibliographystyle{acl_natbib}

\appendix

\clearpage
\section{Training Settings} \label{sec:train}
In our experiments all the models of additive learning, Non-STG and STG, are used the same architecture and hyper-parameters (except whether to use pre-training) for training as described in Table ~\ref{tab:hyper}. 
We found that pre-training the addtional layer of Non-STG-RL with MLE helps the performance improvements. 
On the other hand, STG without pre-training shows better performances. We use the training data for each topic of the task of Data-to-Text as their validation data.

\section{Additonal Study} \label{sec:study}
\subsection{STG-MLE} \label{sec:stg_mle}
Here, we evaluate the MLE version of STG (denoted as STG-MLE) which is trained by MLE for the mixture policy $\pi_{h}(\cdot) = \pi_s(i_t=0)\pi_{LM}(\cdot) + \pi_s(i_t=1)\pi_a(\cdot)$ similar to \textit{copy mechanism} \citep{gu-etal-2016-incorporating}.
In few-shot training, the explicit use of PLM logits can efficiently reduce the fine-tuning loss especially when the adapter is light since the adapter can focus only on the task-relevant part in generation. 
STG-RL\footnote{We add "-RL" to the STG to distinguish with STG-MLE in this context.} learns to do this naturally by stochastic policy sampling if the policy selector is initialized to perform uniform sampling. 
On the other hand, STG-MLE can be easily collapsed to select only a task-specific policy (i.e. $i_t=1$). 
This is because the gradient flows the additional model only and, unlike STG-RL, there is no chance to exploit diverse paths during training in the teacher forcing manner.
As shown in Figure~\ref{fig:curve}, the score of STG-MLE starts from the same point of STG-RL but it collapsed to Non-STG-MLE.

\subsection{Learning Curve} \label{sec:curve}
\begin{figure*}[h]
    \centering
    \includegraphics[width=1\textwidth]{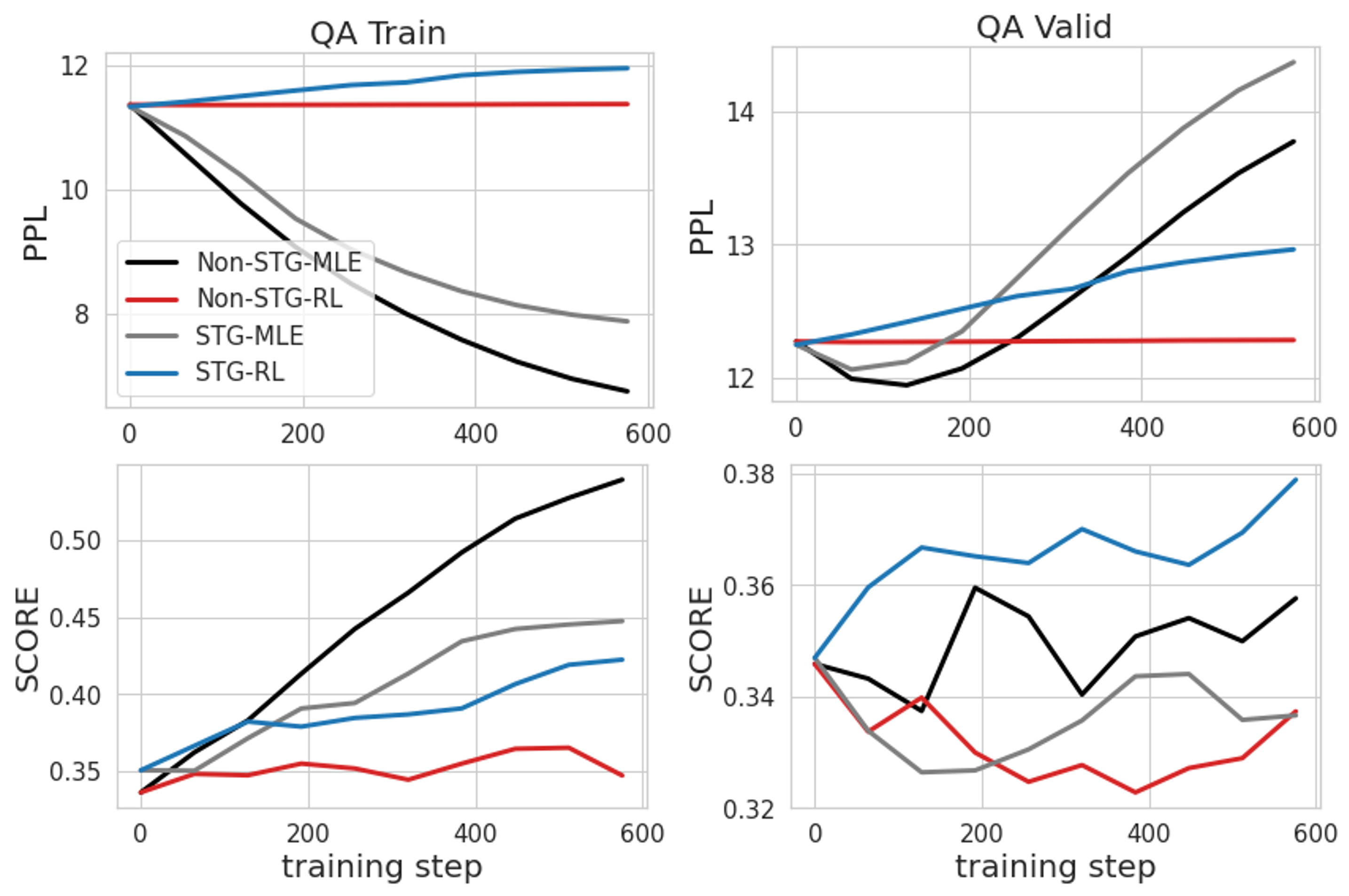}
    \includegraphics[width=1\textwidth]{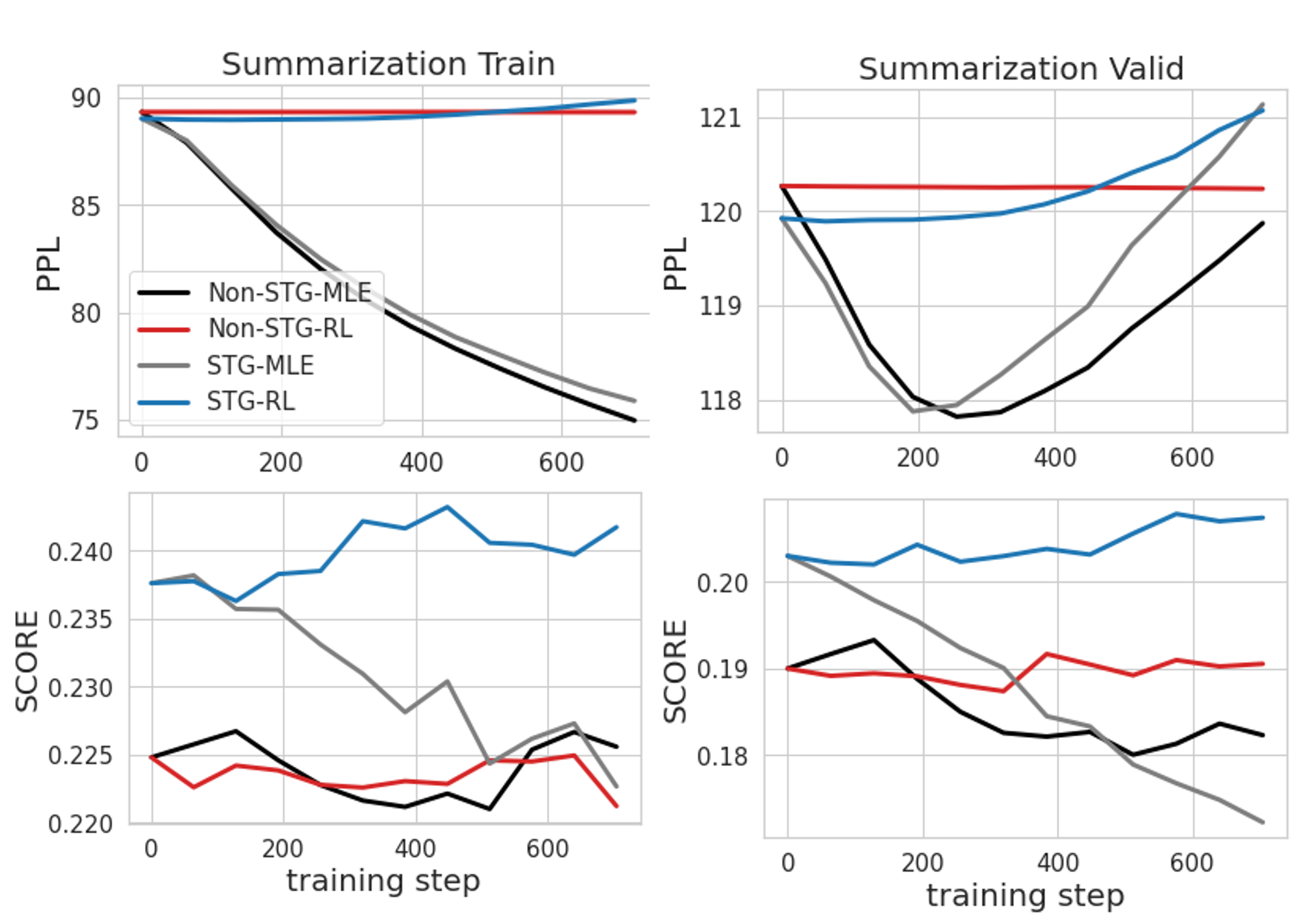}
    \caption{The learning curve. The Perplexity (PPL) and Score of each task (e.g. Rouge-L for Summarization) are measured on the $0.5\%$ few-shot train set and the valid set.}
    \label{fig:curve}
\end{figure*}

It is well known that the RL-tuning resolves the exposure bias of MLE-tuning. We can expect that an additive learner of MLE would be affected by the exposure bias as well, and the RL objective for additive learning resolves it. 
Here, we present some learning curves\footnote{The curve for Data-to-Text is not presented since there is no actual validation set.} obtained from training in our experiments.
As shown in Figure~\ref{fig:curve}, the learners of MLE seem to have overfitting (in terms of Perplexity, PPL) and exposure bias (in terms of Score). 
On the other hand, the learners of RL were less effected by the problems.
We can find that the STGs (denoted STG-RL) are superior to the others from the perspective of the score.

\subsection{Effectiveness of Selector} \label{sec:selector}
\begin{figure*}[h]
    \centering
    \includegraphics[width=1\textwidth]{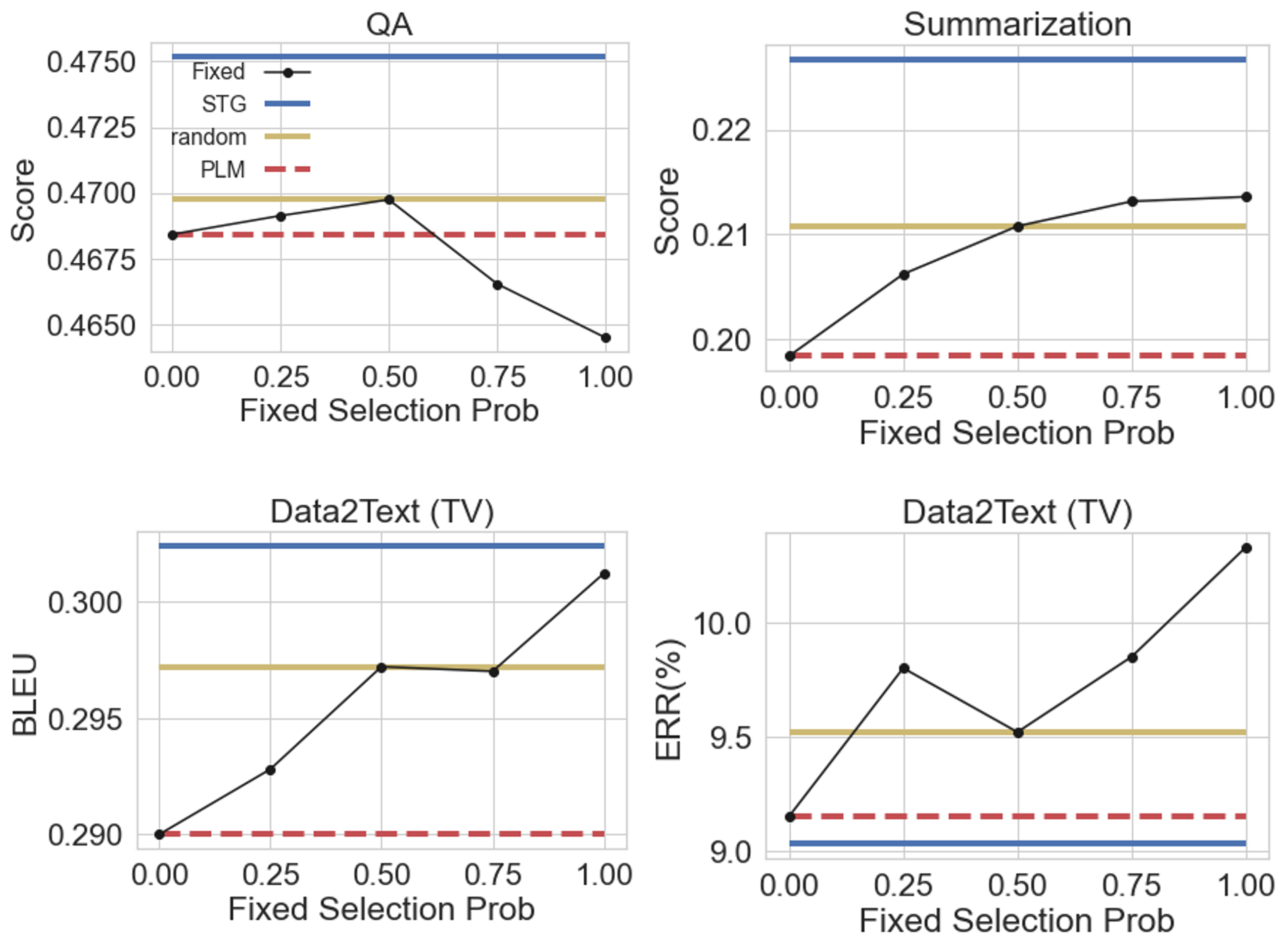}
    \caption{Dynamic selection vs Fixed selection.}
    \label{fig:fixed_inj}
\end{figure*}

Here, we investigate the effectiveness of the selector $\pi_s$ of the STG. We compare \textit{Fixed Selection} against the \textit{Dynamic selection}. In the fixed selection, the probability of selecting the PLM's policy $\pi_{LM}$ is fixed to $\pi_{s}(i_t=0|s_t) = 1 - \pi_{s}(i_t=1|s_t)$. We measure the performance with respect to $\pi_{s}(i_t=1|s_t) = c$ where $c$ is a constant. The selection will be uniformly random when $c=0.5$, and when $c=0$, the performance will be equivalent to the performance of the PLM without additive learning. 
Figure~\ref{fig:fixed_inj} shows that the input-dependent dynamic selection by our STG outperforms the fixed selection with any $c$. We can find that how $\pi_s$ works for each task. For instance, in QA task, the first few tokens of an answer may decide the quality of generation (i.e. "yes" or "no" in binary QA). 
Therefore, an optimal strategy of the STG might be producing the first few tokens sampled from the task-specific $\pi_{a}$ and the remaining tokens from the PLM $\pi_{LM}$. 
The curve supports this interpretation since the score is decreased as $c$ is close to 1. 
Our STG learns such a strategy as shown from the generated answers in Table \ref{tab:qa_exam} and \ref{tab:qa_exam2}.
In Data-to-Text, the BLEU score is increased as $c$ is close to 1 while the ERR score is decreased. This fact supports the results of NE($mix$) models as discussed in Section~\ref{section:result}. The $\pi_s$ learns to balance between the BLEU and ERR.

\section{Generated Sentence Examples} \label{sec:gen_sents}
Here, we show generated sentence examples for each task (see Table \ref{tab:tod_exam} for Data-to-Text, Table~\ref{tab:qa_exam} and Table~\ref{tab:qa_exam2} for Question Answering and Table~\ref{tab:summ_exam} and Table~\ref{tab:summ_exam2} for Summarization.). The tokens sampled from the task-specific policy $\pi_a$ are presented in red.



\clearpage 

\input{./exp_data/app_hyper}

\clearpage 
\input{./exp_data/tod_example}
\input{./exp_data/qa_app_example}
\input{./exp_data/summ_example}
\end{document}

%% file: exp_data/qa_example_intro.tex
\bgroup
\def\arraystretch{1.5}%
\begin{table*}
\begin{tabularx}{\textwidth}{l|X}
\hline
\textbf{Passage} & three types of conflicts are : 1. intrapersonal conflicts , 2. interpersonal conflicts and 3. unconscious conflicts . the word conflict has been derived from a latin word “conflicts” which means “strike two things at the same time” . conflict is \textsuperscript{1)}\textbf{an opposition or a tug-of-war between contradictory impulses .} according to colman "a conflict is \textsuperscript{2)}\textbf{the anticipated frustration entailed in the choice of either alternative}". 
\\ \hdashline
\textbf{Query} & conflict definition psychology 
\\ \hdashline
\textbf{Ground-truth} & the anticipated frustration entailed in the choice of either alternative.
\\
\textbf{PLM} & conflict definition psychology.
\\
\textbf{Adapter} & conflict is an opposition or a tug-of-war between contradictory impulses.
\\
\textbf{PLM with Condition} & \underline{the meaning of conflict is} \textsubscript{(provided condition)} the anticipated frustration entailed in the choice of either alternative.
\\
\textbf{Proposed STG}     & \textbf{\textcolor{red}{conflict is}} the anticipated frustration entailed in the choice of either alternative.
\\ \hline
\end{tabularx}
\caption{\label{tab:qa_example_intro}Generated answers from an instance of MS-MARCO QA dataset. Two definitions about \textit{conflict} are presented in bold text in the passage. The answers are sampled from the models trained on $0.5\%$ few-shot subset data. The proposed selective token generation (STG) produces the first two words (highlighted in \textbf{\textcolor{red}{red}}) by the task-specific adapter while the others by the PLM.} 
\end{table*}
\egroup

%% file: exp_data/tod.tex
\bgroup
\def\arraystretch{1.15}%
\begin{table*}[t]
\small
\centering
\begin{tabular}{lcccccccc}
\hline
\multicolumn{1}{c}{} &
  \multicolumn{2}{c}{Restaurant} &
  \multicolumn{2}{c}{Hotel} &
  \multicolumn{2}{c}{TV} &
  \multicolumn{2}{c}{Laptop} \\
\multicolumn{1}{c}{Model} &
  \multicolumn{1}{c}{BLEU $\uparrow$} &
  \multicolumn{1}{c}{ERR $\downarrow$} &
  \multicolumn{1}{c}{BLEU $\uparrow$} &
  \multicolumn{1}{c}{ERR $\downarrow$} &
  \multicolumn{1}{c}{BLEU $\uparrow$} &
  \multicolumn{1}{c}{ERR $\downarrow$} &
  \multicolumn{1}{c}{BLEU $\uparrow$} &
  \multicolumn{1}{c}{ERR $\downarrow$} \\ \hline
PLM         & 19.42 & 12.57 & 35.84 & 13.74 & 29.0 & 9.15 & 28.27 & 9.31 \\ \hline
Non-STG-MLE & 17.21 & 15.87 & 28.42 & 12.64 & 29.83 & 10.05 & 26.76 & 10.52 \\
Non-STG-RL  & 18.01 & 11.98 & 36.72 & 12.64 & 28.66 & 9.19 & 28.59 & 9.21 \\ \hline
NE($max$)-MLE & 14.12 & 15.27 & 31.32 & 14.29 & 28.23 & 10.21 &  26.93 & 10.02 \\
NE($mix$)-MLE & \textbf{25.27} & 14.97 & 37.13 & 15.93 & \textbf{32.85} & 16.31 & \textbf{32.91} & 14.77 \\
NE($max$)-RL & 15.2 & 11.68 & 32.68 & 16.48 & 28.91 & 9.24 & 28.66 & 9.51  \\
NE($mix$)-RL & 24.1  & 19.16 & 38.07 & 18.68 & 32.84 & 18.06 & 32.53 & 17.14  \\ \hline
STG  & 21.28 & \textbf{10.78} & \textbf{38.09} & \textbf{11.54} & 30.24 & \textbf{9.03} & 30.41 & \textbf{8.91} \\
\hline
\end{tabular}
\caption{\label{tab:tod}Data-to-Text performance on FewShotWOZ dataset. }
\end{table*}
\egroup

%% file: exp_data/qa_1.tex
\bgroup
\def\arraystretch{1.15}%
\begin{table*}[t]
\footnotesize
\centering
\begin{minipage}[t]{0.305\linewidth}
    \subfloat{
    \scalebox{1}{
        \begin{tabular}{lcc}
        \hline
        & \multicolumn{2}{c}{$50 \text{ shot}$} 
        \\
        \multicolumn{1}{c}{Model} & BLEU & R-L
        \\ \hline
        PLM & 19.99 & 29.01
        \\ \hline
        Non-STG-MLE & 27.46 & 35.08
        \\
        Non-STG-RL & 20.07 & 28.94
        \\ \hline
        NE($max$)-MLE & 27.21 & 34.95
        \\
        NE($mix$)-MLE & 26.97 & 35.1
        \\
        NE($max$)-RL & 20.05 & 28.9
        \\
        NE($mix$)-RL & 20.69 & 29.62
        \\ \hline
        STG & \textbf{33.33} & \textbf{39.59}
        \\ \hline
        \end{tabular}
    }}
\end{minipage}
\begin{minipage}[t]{0.15\linewidth}
    \subfloat{
    \scalebox{1}{
        \begin{tabular}{cc}
        \hline
        \multicolumn{2}{c}{$100 \text{ shot}$} 
        \\
        BLEU & R-L
        \\ \hline
        34.93 & 41.27
        \\ \hline
        34.08 & 40.93
        \\
        35.08 & 41.28
        \\ \hline
        34.76 & 41.87
        \\
        35.31 & 41.82
        \\
        35.0 & 41.16
        \\
        35.11 & 41.33
        \\ \hline
        \textbf{36.3} & \textbf{43.24}
        \\ \hline
        \end{tabular}
    }}
\end{minipage}
\begin{minipage}[t]{0.15\linewidth}
    \subfloat{
    \scalebox{1}{
        \begin{tabular}{cc}
        \hline
        \multicolumn{2}{c}{$500 \text{ shot}$} 
        \\
        BLEU & R-L
        \\ \hline
        35.64 & 43.10
        \\ \hline
        34.53 & 43.08
        \\
        35.08 & 42.78
        \\ \hline
        34.69 & 43.93
        \\
        36.26 & 44.43
        \\
        35.14 & 42.94
        \\
        35.93 & 43.52
        \\ \hline
        \textbf{37.37} & \textbf{44.53}
        \\ \hline
        \end{tabular}
    }}
\end{minipage}
\begin{minipage}[t]{0.15\linewidth}
    \subfloat{
    \scalebox{1}{
        \begin{tabular}{cc}
        \hline
        \multicolumn{2}{c}{$1,000 \text{ shot}$} 
        \\
        BLEU & R-L
        \\ \hline
        41.49 & 49.76
        \\ \hline
        41.02 & 50.14
        \\
        41.25 & 49.97
        \\ \hline
        41.11 & 50.77
        \\
        42.26 & \textbf{51.14}
        \\
        41.51 & 50.54
        \\
        42.29 & 50.84
        \\ \hline
        \textbf{42.76} & \textbf{51.19}
        \\ \hline
        \end{tabular}
    }}
\end{minipage}
\begin{minipage}[t]{0.15\linewidth}
    \subfloat{
    \scalebox{1}{
        \begin{tabular}{cc}
        \hline
        \multicolumn{2}{c}{$2,000 \text{ shot}$} 
        \\
        BLEU & R-L
        \\ \hline
        47.72 & 56.02
        \\ \hline
        47.85 & 56.81
        \\
        48.00 & 56.83
        \\ \hline
        47.65 & 57.22
        \\
        \textbf{48.44} & \textbf{57.3}
        \\
        47.58 & 57.06
        \\
        48.28 & 57.02
        \\ \hline
        \textbf{48.42} & \textbf{57.3}
        \\ \hline
        \end{tabular}
    }}
\end{minipage}
\caption{\label{tab:qa}\textit{Averaged performances} for Question Answering on various few-shot subset data of MS-MARCO. }
\end{table*}

\begin{figure*}[h]
     \centering
     \scalebox{0.85}{
         \begin{subfigure}[b]{0.48\linewidth}
         \includegraphics[width=\linewidth]{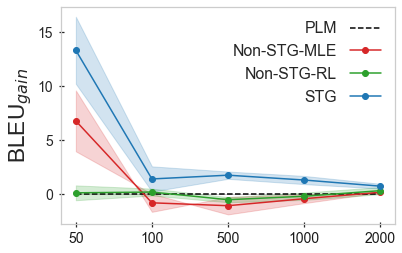}
         \end{subfigure}
    }
     \scalebox{0.85}{
         \begin{subfigure}[b]{0.48\linewidth} \includegraphics[width=\linewidth]{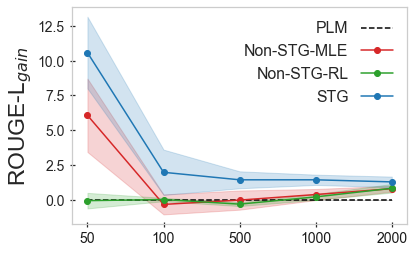}
         \end{subfigure}
     }
    \caption{\textit{Averaged performance gains} against the PLM for Question Answering on various few-shot subset data of MS-MARCO. The x-axis represents the size of the subset data and the shaded area represents a range of standard deviation over 3 randomly sampled subset data with different random seeds. STG provides significantly larger gains compared to Non-STGs on BLEU (Left) and ROUGE-L (Right).}
    \label{fig:QA_gain}
\end{figure*}
\egroup

%% file: exp_data/summ_1.tex
\bgroup
\def\arraystretch{1.15}%
\begin{table*}[h]
\footnotesize
\begin{minipage}[t]{0.298\linewidth}
    \subfloat{
    \scalebox{0.84}{
        \begin{tabular}{lccc}
        \hline
        & \multicolumn{3}{c}{$50 \text{ shot}$}
        \\
        \multicolumn{1}{c}{Model} & R1 & R2 & R-L
        \\ \hline
        PLM & 14.67 & 4.57 & 10.69
        \\ \hline
        Non-STG-MLE & 15.39 & 4.81 & 11.09
        \\
        Non-STG-RL & 15.22 & 4.76 & 11.08
        \\ \hline
        NE($max$)-MLE & 15.52 & 4.89 & 11.24
        \\
        NE($mix$)-MLE & 15.4 & 4.83 & 11.16
        \\
        NE($max$)-RL & 15.14 & 4.73 & 11.02
        \\
        NE($mix$)-RL & 14.95 & 4.67 & 10.89
        \\ \hline
        STG & \textbf{17.4} & \textbf{5.33} & \textbf{12.42}
        \\ \hline
        \end{tabular}
    }}
\end{minipage}
\begin{minipage}[t]{0.167\linewidth}
    \subfloat{
    \scalebox{0.84}{
        \begin{tabular}{ccc}
        \hline
        \multicolumn{3}{c}{$100 \text{ shot}$}
        \\
        R1 & R2 & R-L
        \\ \hline
        16.58 & 5.28 & 12.05
        \\ \hline
        17.09 & 5.41 & 12.3
        \\
        16.55 & 5.25 & 12.0
        \\ \hline
        16.98 & 5.43 & 12.26
        \\
        16.88 & 5.4 & 12.22
        \\
        16.52 & 5.27 & 11.99
        \\
        16.6 & 5.29 & 12.04
        \\ \hline
        \textbf{17.96} & \textbf{5.73} & \textbf{12.94}
        \\ \hline
        \end{tabular}
    }}
\end{minipage}
\begin{minipage}[t]{0.167\linewidth}
    \subfloat{
    \scalebox{0.84}{
        \begin{tabular}{ccc}
        \hline
        \multicolumn{3}{c}{$300 \text{ shot}$}
        \\
        R1 & R2 & R-L
        \\ \hline
        19.38 & 7.08 & 13.74
        \\ \hline
        18.9 & 6.87 & 13.36
        \\
        19.61 & 7.11 & 13.83
        \\ \hline
        19.19 & 7.0 & 13.56
        \\
        19.45 & 7.07 & 13.75
        \\
        19.47 & 7.1 & 13.76
        \\
        19.58 & 7.14 & 13.84
        \\ \hline
        \textbf{23.27} & \textbf{8.32} & \textbf{16.29}
        \\ \hline
        \end{tabular}
    }}
\end{minipage}
\begin{minipage}[t]{0.175\linewidth}
    \subfloat{
    \scalebox{0.84}{
        \begin{tabular}{ccc}
        \hline
        \multicolumn{3}{c}{$1,500 \text{ shot}$}
        \\
        R1 & R2 & R-L
        \\ \hline
        30.19 & 11.27 & 21.21
        \\ \hline
        30.34 & 11.32 & 21.2
        \\
        30.35 & 11.34  & 21.22
        \\ \hline
        30.33 & 11.31 & 21.2
        \\
        30.32 & 11.31 & 21.23
        \\
        30.37 & 11.35 & 21.26
        \\
        30.28 & 11.3 & 21.22
        \\ \hline
        \textbf{30.47} & \textbf{11.37} & \textbf{21.36}
        \\ \hline
        \end{tabular}
    }}
\end{minipage}
\begin{minipage}[t]{0.167\linewidth}
    \subfloat{
    \scalebox{0.84}{
        \begin{tabular}{ccc}
        \hline
        \multicolumn{3}{c}{$3,000 \text{ shot}$}
        \\
        R1 & R2 & R-L
        \\ \hline
        33.05& 12.96 & 23.39
        \\ \hline
        33.19 & 12.98 & 23.39
        \\
        33.22 & 12.99 & 23.4
        \\ \hline
        33.19 & 12.99 & 23.4
        \\
        33.11 & 12.99 & 23.41
        \\
        33.21 & 12.99 & 23.41
        \\
        33.14 & 13.0 & 23.42
        \\ \hline
        \textbf{33.45} & \textbf{13.14} & \textbf{23.66}
        \\ \hline
        \end{tabular}
    }}
\end{minipage}
\caption{\label{tab:summ1}\textit{Averaged performances} for Text Summarization on various few-shot subset data of CNN/DM.}
\end{table*}

\begin{figure*}[h]
     \centering
     \scalebox{0.95}{
         \begin{subfigure}[b]{0.32\linewidth}
         \includegraphics[width=\linewidth]{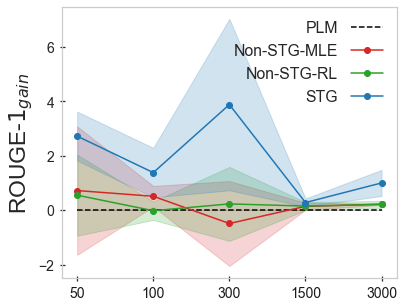}
         \end{subfigure}
    }
     \scalebox{0.95}{
         \begin{subfigure}[b]{0.32\linewidth} \includegraphics[width=\linewidth]{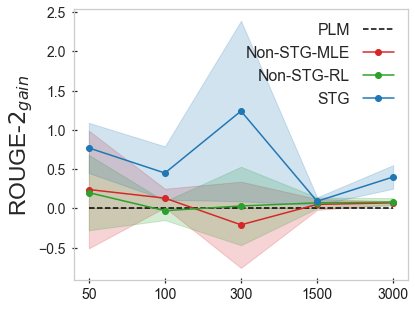}
         \end{subfigure}
     }
      \scalebox{0.95}{
         \begin{subfigure}[b]{0.32\linewidth} \includegraphics[width=\linewidth]{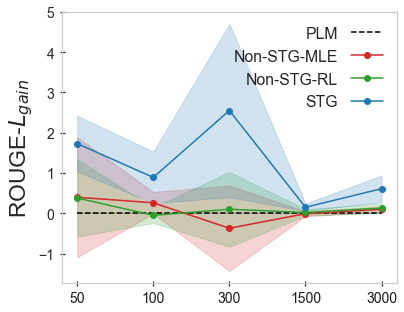}
         \end{subfigure}
     }
    \caption{\textit{Averaged performance gains} against the PLM for Text Summarization on various few-shot subset data of CNN/DM. The x-axis represents the size of the subset data and the shaded area represents a range of standard deviation over 3 randomly sampled subset data with different random seeds. STG provides significantly larger gains compared to Non-STGs on ROUGE-1 (Left), ROUGE-1 (Middle), and ROUGE-L (Right).}
    \label{fig:summ_gain}
\end{figure*}
\egroup

%% file: exp_data/app_hyper.tex
\bgroup
\def\arraystretch{1.15}%
\begin{table*}[h]
\centering
\begin{adjustbox}{width=1.0\textwidth}
\begin{tabular}{llll}
        \multicolumn{1}{c}{\bf Hyper-parameter}  &\multicolumn{1}{c}{\bf Summarization} &\multicolumn{1}{c}{\bf Data-to-Text} &\multicolumn{1}{c}{\bf Question Answering}
        \\ \hline
        Num layer                &\multicolumn{3}{c}{2} \\
        RNN hidden size          &\multicolumn{1}{c}{512}        &\multicolumn{1}{c}{256} &\multicolumn{1}{c}{256} \\
        $\gamma$                 &\multicolumn{3}{c}{1} \\
        Optimizer                &\multicolumn{3}{c}{AdamW with $betas=(0.9, 0.999)$, $eps=10^{-8}$} \\
        Learning rate            &\multicolumn{2}{c}{2e-5} &\multicolumn{1}{c}{5e-5}\\
        Pre-train epoch          &\multicolumn{1}{c}{1}          &\multicolumn{1}{c}{0} &\multicolumn{1}{c}{1}\\
        (Non-STG-RL)             &                               &    \\
        Validation data size              &\multicolumn{1}{c}{500}        &\multicolumn{1}{c}{50} &\multicolumn{1}{c}{500} \\
        Train epochs             &\multicolumn{1}{c}{25 (50 shot), 20 (100 shot), 15 (300 shot),}   
        &\multicolumn{1}{c}{30} 
        &\multicolumn{1}{c}{50 (50 shot), 40 (100 shot), 20 (300 shot),} \\
        & \multicolumn{1}{c}{8 (1,500 shot), 4 (3,000 shot)} & & \multicolumn{1}{c}{10 (1,000 shot), 5 (2,000 shot)}\\
        Batch size               &\multicolumn{1}{c}{16}         &\multicolumn{1}{c}{10}    &\multicolumn{1}{c}{16} \\
\end{tabular}
\end{adjustbox}
\caption{\label{tab:hyper}\textbf{Hyper-parameters used for experiments}}
\end{table*}
\egroup

%% file: exp_data/tod_example.tex
\bgroup
\def\arraystretch{1.5}%
\begin{table*}[h]
\begin{tabularx}{\textwidth}{l|X}

\hline
\textbf{Data} & recommend(name=kratos 31; type=television; powerconsumption=31.4 watt; accessories=remote control and european warranty; color=mat black bezel with mat black frame stand)
\\ \hdashline
\textbf{Ground-truth} & \textcolor{blue}{the kratos 31 is a great television . it only consumes 31.4 watts , comes with a remote control and european warranty and has a stylish mat black bezel with mat black frame stand .}
\\
\textbf{PLM} & the kratos 31 features a mat black bezel with mat black frame stand and european warranty.
\\
\textbf{Non-STG-MLE} & the kratos 31 features a mat black bezel with mat black frame stand and european warranty
\\
\textbf{NE($mix$)-MLE} & \textcolor{red}{the} k\textcolor{red}{ratos 31} television \textcolor{red}{has 31.4 watt power consumption, remote control and european warranty.}
\\
\textbf{STG}     & \textcolor{red}{the kratos 31 television has} 31 \textcolor{red}{watt} power consumption\textcolor{red}{.} \textcolor{red}{it comes with} a remote control \textcolor{red}{and europe}an warranty \textcolor{red}{with a color of mat black bezel with mat black frame stand.}
\\ \hline
\textbf{Data} & inform(name=triton 52; type=television; family=l7; audio=cevo stereo; screensize=47 inch)
\\ \hdashline
\textbf{Ground-truth} & \textcolor{blue}{the triton 52 47 inch screen television from the l7 product family and has cevo stereo audio.}
\\
\textbf{PLM} & the triton 52 television is a 47 inch television in the l7 family and with a cevo stereo.
\\
\textbf{Non-STG-MLE} & the triton 52 television is in the l7 family, it has 47 inches of screen, and uses cevo stereo sound capabilities.
\\
\textbf{NE($mix$)-MLE} & \textcolor{red}{the} tr\textcolor{red}{iton 52} television is \textcolor{red}{in the l7 family}, it \textcolor{red}{has 47 inches of screen, and} a \textcolor{red}{47 euro price,}
\\
\textbf{STG}     & the tr\textcolor{red}{iton 52 television} is a \textcolor{red}{47 inch television in the} l7 family and \textcolor{red}{has a cevo stereo.}
\\ \hline
\textbf{Data} & inform(name=attis 65; type=television; resolution=1080p; pricerange=cheap)
\\ \hdashline
\textbf{Ground-truth} & \textcolor{blue}{the attis 65 television is in the cheap price range and has a nice resolution of 1080p.}
\\
\textbf{PLM} & the attis 65 television in the cheap price range has a 1080p resolution and a pricerange of cheap.
\\
\textbf{Non-STG-MLE} & the attis 65 television has a 1080p resolution, is in the cheap price range and has a nicam stereo.
\\
\textbf{NE($mix$)-MLE} & \textcolor{red}{the attis 65} television is a \textcolor{red}{1080p} television \textcolor{red}{in the cheap price range} and in \textcolor{red}{the} cheap price \textcolor{red}{range} with \textcolor{red}{a} nic\textcolor{red}{am} stereo\textcolor{red}{.}
\\
\textbf{STG}     & \textcolor{red}{the attis 65} television is \textcolor{red}{in} the cheap \textcolor{red}{price range and has} a \textcolor{red}{1080}p \textcolor{red}{resolution.}
\\ \hline
\end{tabularx}
\caption{\label{tab:tod_exam}Data-to-Text examples from TV of FewShotWOZ dataset.}
\end{table*}
\egroup

%% file: exp_data/qa_app_example.tex
\bgroup
\def\arraystretch{1.5}%
\begin{table*}[h]
\begin{tabularx}{\textwidth}{l|X}
\hline
\textbf{Passage} & zygote : this cell results from the union of an oocyte and a sperm . a zygote is the beginning of a new human being ( i.e. , an embryo ) . the expression fertilized ovum refers to a secondary oocyte that is impregnated by a sperm ; when fertilization is complete , the oocyte becomes a zygote.10 ( emphasis added . )
\\ \hdashline
\textbf{Query} & what is a human zygote
\\ \hdashline
\textbf{Ground-truth}      & \textcolor{blue}{the beginning of a new human being .}
\\
\textbf{PLM} & the expression fertilized ovum refers to a secondary oocyte that is impregnated by a sperm ;
\\
\textbf{NE($mix$)-MLE} & the expression fertilized ovum refers to a secondary oocyte that is impregnated by a sperm ; when fertilization is complete , the oocyte becomes a zygote .
\\
\textbf{NE($mix$)-RL} & the expression fertilized ovum refers to a secondary oocyte that is impregnated by a sperm ; when fertilization is complete , the oocyte becomes a zygote .
\\
\textbf{STG}     & \textcolor{red}{it} is the beginning of a new human being .
\\ \hline
\textbf{Passage} & sometimes the terms reactive airway disease and asthma are used interchangeably , but they are not the same thing . often , the term reactive airway disease is used when asthma is suspected , but not yet confirmed . reactive airway disease in children is a general term that does n't indicate a specific diagnosis . it may be used to describe a history of coughing , wheezing or shortness of breath triggered by infection . these signs and symptoms may or may not be caused by asthma .
\\ \hdashline
\textbf{Query} & symptoms of reactive airway disease in adults
\\ \hdashline
\textbf{Ground-truth}      & \textcolor{blue}{coughing , wheezing or shortness of breath triggered by infection .}
\\
\textbf{PLM} & symptoms of reactive airway disease in children is a general term that does n\'t indicate a specific diagnosis .
\\
\textbf{NE($mix$)-MLE} & symptoms of reactive airway disease in adults are a general term that does n\'t indicate a specific diagnosis .
\\
\textbf{NE($mix$)-RL} & yes , reactive airway disease in children is a general term that does n\'t indicate a specific diagnosis .
\\
\textbf{STG}     & \textcolor{red}{c}oughing , wheezing or shortness of breath triggered by infection .
\\ \hline
\textbf{Passage} & sulfate is an anion , composed of a sulfur atom and four oxygen atoms . the overall charge of this anion is -2. the molecular formula for sulfate is so 4 2-. there , two oxygen atoms are bonded to the sulfur atom through double bonds and other two are bonded to sulfur through single bonds . the oxidation state of sulfur in sulfate is +6 .
\\ \hdashline
\textbf{Query} & is sulfate sulfur and oxygen bonded
\\ \hdashline
\textbf{Ground-truth}      & \textcolor{blue}{yes , sulfate is sulfur and oxygen bonded .}
\\
\textbf{PLM} & no answer present .
\\
\textbf{NE($mix$)-MLE} & yes , sulfate is sulfur and oxygen bonded to the sulfur atom through double bonds and other two are bonded to sulfur through single bonds .
\\
\textbf{NE($mix$)-RL} & yes , sulfate sulfur and oxygen bonded to the sulfur atom through double bonds and other two are bonded to sulfur through single bonds .
\\
\textbf{STG}     & \textcolor{red}{yes} , sulfate is sulfur and oxygen bonded to each other through double bonds and other two are bonded to sulfur through single bonds .
\\ \hline
\end{tabularx}
\caption{\label{tab:qa_exam}Question Answering examples.}
\end{table*}
\egroup

\bgroup
\def\arraystretch{1.5}%
\begin{table*}[h]
\begin{tabularx}{\textwidth}{l|X}
\hline
\textbf{Passage} & gryffindor house is one of the four houses at hogwart into which the students are sorted when they start their first year . gryffindor is the hogwarts house of founder godric gryffindor . gryffindor house places high value on traits such as courage , chivalry , and nerve . during harry ’ s time at hogwarts , the head of gryffindor house was minerva [ … ] 
\\ \hdashline
\textbf{Query} & what kind of house is a gryffindor
\\ \hdashline
\textbf{Ground-truth}      & \textcolor{blue}{gryffindor house is one of the four houses at hogwart into which the students are sorted when they start their first year .}
\\
\textbf{PLM} & gryffindor house is one of the four houses at hogwart into which the students are sorted when they start their first year .
\\
\textbf{NE($mix$)-MLE} & no answer present .
\\
\textbf{NE($mix$)-RL} & no answer present .
\\
\textbf{STG}     & \textcolor{red}{gry}ffindor house is one of the four houses at hogwarts into which the students are sorted when they start their first year .
\\ \hline
\textbf{Passage} & the median home value in louisville is \$ 119,500 . louisville home values have gone up 1.8 \% over the past year and zillow predicts they will rise 3.0 \% within the next year . the median list price per square foot in louisville is \$ 125 . the median price of homes currently listed in louisville is \$ 180,000 . the median rent price in louisville is \$ 1,050 . foreclosures will be a factor impacting home values in the next several years . in louisville 0.1 homes are foreclosed ( per 10,000 ) .
\\ \hdashline
\textbf{Query} & average per therm price in louisville ky
\\ \hdashline
\textbf{Ground-truth}      & \textcolor{blue}{the median home value in louisville is \$ 119,500 .}
\\
\textbf{PLM} & the average per therm price in louisville is \$ 119,500 .
\\
\textbf{NE($mix$)-MLE} & \$ 125 per square foot
\\
\textbf{NE($mix$)-RL} & the average per square foot in louisville is \$ 125 .
\\
\textbf{STG} & \textcolor{red}{the median home} value in louisville is \$ 119,500 .
\\ \hline
\textbf{Passage} & the brain is made up of more than just nerve cells ( neurons ) . although there are about 100 billion neurons in the brain , there may be about 10 to 50 times that many glial cells in the brain . neurons have two processes called axons and dendrites .... glial cells have only one . 2 neurons can generate action potentials ... glial cells can not . 3 however , glial cells do have a resting potential . 4 neurons have synapses that use neurotransmitters ... glial cells do not have chemical synapses .
\\ \hdashline
\textbf{Query} & number of glial cells vs neurons
\\ \hdashline
\textbf{Ground-truth}      & \textcolor{blue}{10 to 50 times .}
\\
\textbf{PLM} & no answer present .
\\
\textbf{NE($mix$)-MLE} & \$ 10 to 50 times that many glial cells in the brain .
\\
\textbf{NE($mix$)-RL} & no answer present .
\\
\textbf{STG} & \textcolor{red}{10} to 50 times
\\ \hline
\end{tabularx}
\caption{\label{tab:qa_exam2}Question Answering examples.}
\end{table*}
\egroup

%% file: exp_data/summ_example.tex
\bgroup
\def\arraystretch{1.25}%
\begin{table*}[h]
\begin{tabularx}{\textwidth}{l|X}
\hline
\textbf{Article} & -lrb- cnn -rrb- -- three runners died sunday during the detroit free press/flagstar marathon in detroit, michigan, police told cnn. an emt vehicle is at the scene sunday in detroit after three runners collapsed at a marathon. all three deaths occurred between 9 and 9:20 a.m. et, second deputy chief john roach said. a man in his 60s fell and hit his head, roach said. the cause of the fall was unknown. the man was transported to detroit receiving hospital, where he was pronounced dead. two other men, ages 36 and 26, also collapsed during the race and were pronounced dead at the hospital, roach said. all three collapsed near the end of the race, [...]
\\ \hdashline
\textbf{Ground-truth}      & \textcolor{blue}{second deputy chief john roach : all three deaths occurred between 9 and 9:20 a.m. man in his 60s fell hit his head ; two men others , ages 36 and 26 , collapsed . race was detroit free press/flagstar marathon in detroit , michigan .}
\\
\textbf{PLM} & three runners collapsed at a marathon in detroit , police say . the cause of the fall is unknown .
\\
\textbf{Non-STG-MLE}     & three runners collapsed at a marathon sunday , police say . the cause of the fall is unknown , police say .
\\
\textbf{Non-STG-RL}     & three runners collapsed at a marathon sunday , police say . the cause of the fall is unknown , police say .
\\
\textbf{STG}     & \textcolor{red}{three runners collapsed at a marathon in detroit , michigan . all three deaths occurred between} 9 and 9:20 a.m. et . a man in his 60s fell and hit his head , police say .
 
\\ \hline

\textbf{Article} & london, england -lrb- cnn -rrb- -- up to 1,000 human rights campaigners demonstrated saturday in front of no. 10 downing street, the official residence of british prime minister gordon brown, calling on the british government to demand that full democracy be restored in pakistan. jemima khan, center, ex-wife for former pakistani cricket star imran khan, joins protesters in london. protesters waved placards and chanted in support of the resignation of pakistani president pervez musharraf, a week after he imposed a state of emergency in the country. the crowd of demonstrators massed behind barriers and included jemima khan, the ex-wife of former pakistani cricket star turned politician imran khan. the demonstrators carried placards saying `` free the innocent '' and `` end musharraf's regime '' and waved pakistani flags. imran khan, who heads the the movement for justice party, has been under house arrest since the emergency declaration. his ex-wife delivered a petition to a doorman at downing street, [...]          
\\ \hdashline
\textbf{Ground-truth}      & \textcolor{blue}{human rights campaigners demonstrate in front of no . 10 downing street . protests urged uk government to demand full democracy restored in pakistan . cricketer turned politician imran khan \'s ex wife jemima among protesters .}  
\\
\textbf{PLM} & pakistani president pervez musharraf has been under house arrest since the emergency declaration . his ex-wife delivered a petition to a doorman at downing street . 
\\
\textbf{Non-STG-MLE}     & pakistani president pervez musharraf has imposed a state of emergency in the country . he has been under house arrest since the emergency declaration . 
\\
\textbf{Non-STG-RL}     & pakistani president pervez musharraf has imposed a state of emergency in the country . he has imposed a state of emergency in the country since last week .         
\\
\textbf{STG} & \textcolor{red}{the official residence of british prime minister gordon brown . hundreds} of protesters demonstrate in front of no . 10 downing street . the petition calls for full democracy in pakistan .  
\\ \hline
\end{tabularx}
\caption{\label{tab:summ_exam}Text Summarization examples.}
\end{table*}
\egroup

\bgroup
\def\arraystretch{1.25}%
\begin{table*}[h]
\begin{tabularx}{\textwidth}{l|X}
\hline
\textbf{Article} & -lrb- cnn -rrb- spoiler alert! it's not just women getting cloned. that was the big twist at the end of `` orphan black's '' second season. the kickoff to the new season leads the list of six things to watch in the week ahead. 1. `` orphan black, '' 9 p.m. et, saturday, april 18, bbc america. the cloning cult sci-fi series remains one of the most critically acclaimed shows on tv, thanks in large part to the performance of tatiana maslany, who has taken on at least six roles on the show so far, [...] 2. `` turn : washington's spies, '' 9 p.m. et, monday, amc. the series about spies in the early days of the revolutionary war returns with a new subtitle, `` washington's spies, '' and a new monday night time slot. [...] 3. `` game of thrones, '' 9 p.m. et, sunday, hbo. the world of westeros returns for a fifth season in one of the biggest season premieres of the year. click here for more on what to expect. [...]
\\ \hdashline
\textbf{Ground-truth}      & \textcolor{blue}{critically acclaimed series `` orphan black `` returns . `` turn : washington \'s spies `` starts a second season . `` game of thrones `` is back for season five .} 
\\
\textbf{PLM} & series returns with a new subtitle , `` washington \'s spies , `` and a new monday night time slot . series star jamie bell told cnn what we can expect in the second season . 
\\
\textbf{Non-STG-MLE}     & `` orphan black `` returns with a new subtitle and a new monday night time slot . series star tatiana maslany is blown away by the response to the series so far .     
\\
\textbf{Non-STG-RL}     & `` orphan black `` returns with a new subtitle and a new monday night time slot . series star jamie bell says the series will bring more recognition to the culper spy ring .         
\\
\textbf{STG}     & \textcolor{red}{the new season of `` orphan} black `` premieres on saturday the 18th . `` game of thrones `` returns for a fifth season .                       
\\ \hline

\textbf{Article} & energy drinks can pose a real problem for children, according to a new study from the american heart association. researchers found that 40 \% of the 5,156 calls to poison centers for `` energy drink exposure '' involved children under age 6. in most of the cases, the parents didn't know the children had gotten hold of an energy drink. many of the calls reported the children were experiencing serious side effects, such as an abnormal heart rhythm, or they were having a seizure. the study is being presented at the american heart association's scientific sessions this week. study : caffeine affects teen boys more. study author dr. steven lipshultz has handled cases involving children who became sick after consuming energy drinks. while studies about the impact of caffeine on children are limited, lipshultz, [...]
\\ \hdashline
\textbf{Ground-truth}      & \textcolor{blue}{40 \% of `` energy drink exposure `` calls to poison control centers are for children under 6. study authors believe the number is under-counted . the fda does not set a caffeine limit for children .}    
\\
\textbf{PLM} & drug calls to poison centers for `` energy drink exposure `` in most cases , the parents did n't know the children had gotten hold of an energy drink . many of the calls reported the children were experiencing serious side effects , such as an abnormal heart rhythm . 
\\
\textbf{Non-STG-MLE}     & drug calls to poison centers for `` energy drink exposure `` in most cases , the parents did n't know the children had gotten hold of an energy drink . many of the calls reported the children were experiencing serious side effects , such as an abnormal heart rhythm .
\\
\textbf{Non-STG-RL}     & `` this is a very concerning finding , `` dr. laurence sperling says . the american academy of pediatrics recommends children consume no caffeine .
\\
\textbf{STG} & \textcolor{red}{drug calls to poison centers for `` energy drink exposure `` nearly 40 \% of calls to poison centers for `` energy drink exposure ``} involved children under age 6. study : caffeine affects teens more .  
\\ \hline
\end{tabularx}
\caption{\label{tab:summ_exam2}Text Summarization examples.}
\end{table*}
\egroup

%% file: acl_latex.bbl
\begin{thebibliography}{46}
\expandafter\ifx\csname natexlab\endcsname\relax\def\natexlab#1{#1}\fi

\bibitem[{Alayrac et~al.(2022)Alayrac, Donahue, Luc, Miech, Barr, Hasson, Lenc,
  Mensch, Millican, Reynolds, Ring, Rutherford, Cabi, Han, Gong, Samangooei,
  Monteiro, Menick, Borgeaud, Brock, Nematzadeh, Sharifzadeh, Binkowski,
  Barreira, Vinyals, Zisserman, and Simonyan}]{flamingo}
Jean-Baptiste Alayrac, Jeff Donahue, Pauline Luc, Antoine Miech, Iain Barr,
  Yana Hasson, Karel Lenc, Arthur Mensch, Katie Millican, Malcolm Reynolds,
  Roman Ring, Eliza Rutherford, Serkan Cabi, Tengda Han, Zhitao Gong, Sina
  Samangooei, Marianne Monteiro, Jacob Menick, Sebastian Borgeaud, Andrew
  Brock, Aida Nematzadeh, Sahand Sharifzadeh, Mikolaj Binkowski, Ricardo
  Barreira, Oriol Vinyals, Andrew Zisserman, and Karen Simonyan. 2022.
\newblock \href {https://doi.org/10.48550/ARXIV.2204.14198} {Flamingo: a visual
  language model for few-shot learning}.

\bibitem[{Bahdanau et~al.(2017)Bahdanau, Brakel, Xu, Goyal, Lowe, Pineau,
  Courville, and Bengio}]{bahdanau2017actor}
Dzmitry Bahdanau, Philemon Brakel, Kelvin Xu, Anirudh Goyal, Ryan Lowe, Joelle
  Pineau, Aaron Courville, and Yoshua Bengio. 2017.
\newblock \href {https://arxiv.org/abs/1607.07086} {An actor-critic algorithm
  for sequence prediction}.
\newblock \emph{International Conference on Learning Representations}.

\bibitem[{Brown et~al.(2020)Brown, Mann, Ryder, Subbiah, Kaplan, Dhariwal,
  Neelakantan, Shyam, Sastry, Askell, Agarwal, Herbert-Voss, Krueger, Henighan,
  Child, Ramesh, Ziegler, Wu, Winter, Hesse, Chen, Sigler, Litwin, Gray, Chess,
  Clark, Berner, McCandlish, Radford, Sutskever, and Amodei}]{Brown2020gpt3}
Tom~B. Brown, Benjamin Mann, Nick Ryder, Melanie Subbiah, Jared Kaplan,
  Prafulla Dhariwal, Arvind Neelakantan, Pranav Shyam, Girish Sastry, Amanda
  Askell, Sandhini Agarwal, Ariel Herbert-Voss, Gretchen Krueger, Tom Henighan,
  Rewon Child, Aditya Ramesh, Daniel~M. Ziegler, Jeffrey Wu, Clemens Winter,
  Christopher Hesse, Mark Chen, Eric Sigler, Mateusz Litwin, Scott Gray,
  Benjamin Chess, Jack Clark, Christopher Berner, Sam McCandlish, Alec Radford,
  Ilya Sutskever, and Dario Amodei. 2020.
\newblock Language models are few-shot learners.
\newblock \emph{arXiv preprint arXiv:2005.14165}.

\bibitem[{Chandak et~al.(2019)Chandak, Theocharous, Kostas, Jordan, and
  Thomas}]{chandak2019learning}
Yash Chandak, Georgios Theocharous, James Kostas, Scott Jordan, and Philip
  Thomas. 2019.
\newblock Learning action representations for reinforcement learning.
\newblock In \emph{International Conference on Machine Learning}, pages
  941--950. PMLR.

\bibitem[{Chen et~al.(2020)Chen, Eavani, Chen, Liu, and
  Wang}]{chen-etal-2020-shot}
Zhiyu Chen, Harini Eavani, Wenhu Chen, Yinyin Liu, and William~Yang Wang. 2020.
\newblock \href {https://doi.org/10.18653/v1/2020.acl-main.18} {Few-shot {NLG}
  with pre-trained language model}.
\newblock In \emph{Proceedings of the 58th Annual Meeting of the Association
  for Computational Linguistics}, pages 183--190, Online. Association for
  Computational Linguistics.

\bibitem[{Dathathri et~al.(2020)Dathathri, Madotto, Lan, Hung, Frank, Molino,
  Yosinski, and Liu}]{Dathathri2020Plug}
Sumanth Dathathri, Andrea Madotto, Janice Lan, Jane Hung, Eric Frank, Piero
  Molino, Jason Yosinski, and Rosanne Liu. 2020.
\newblock \href {https://openreview.net/forum?id=H1edEyBKDS} {Plug and play
  language models: A simple approach to controlled text generation}.
\newblock In \emph{International Conference on Learning Representations}.

\bibitem[{Devlin et~al.(2019)Devlin, Chang, Lee, and
  Toutanova}]{devlin-etal-2019-bert}
Jacob Devlin, Ming-Wei Chang, Kenton Lee, and Kristina Toutanova. 2019.
\newblock \href {https://doi.org/10.18653/v1/N19-1423} {{BERT}: Pre-training of
  deep bidirectional transformers for language understanding}.
\newblock In \emph{Proceedings of the 2019 Conference of the North {A}merican
  Chapter of the Association for Computational Linguistics: Human Language
  Technologies, Volume 1 (Long and Short Papers)}, pages 4171--4186,
  Minneapolis, Minnesota. Association for Computational Linguistics.

\bibitem[{Ding and Soricut(2017)}]{Ding2017}
Nan Ding and Radu Soricut. 2017.
\newblock Cold-start reinforcement learning with softmax policy gradient.
\newblock \emph{arXiv preprint arXiv:1709.09346}.

\bibitem[{Dulac-Arnold et~al.(2015)Dulac-Arnold, Evans, van Hasselt, Sunehag,
  Lillicrap, Hunt, Mann, Weber, Degris, and Coppin}]{dulac2015deep}
Gabriel Dulac-Arnold, Richard Evans, Hado van Hasselt, Peter Sunehag, Timothy
  Lillicrap, Jonathan Hunt, Timothy Mann, Theophane Weber, Thomas Degris, and
  Ben Coppin. 2015.
\newblock Deep reinforcement learning in large discrete action spaces.
\newblock \emph{arXiv preprint arXiv:1512.07679}.

\bibitem[{Elsahar et~al.(2018)Elsahar, Gravier, and Laforest}]{Elsahar2018}
Hady Elsahar, Christophe Gravier, and Frederique Laforest. 2018.
\newblock Zero-shot question generation from knowledge graphs for unseen
  predicates and entity types.
\newblock \emph{arXiv preprint arXiv:1802.06842}.

\bibitem[{Even-Dar et~al.(2003)Even-Dar, Mannor, and Mansour}]{even2003action}
Eyal Even-Dar, Shie Mannor, and Yishay Mansour. 2003.
\newblock Action elimination and stopping conditions for reinforcement
  learning.
\newblock In \emph{Proceedings of the 20th International Conference on Machine
  Learning (ICML-03)}, pages 162--169.

\bibitem[{Fedus et~al.(2018)Fedus, Goodfellow, and Dai}]{fedus2018maskgan}
William Fedus, Ian Goodfellow, and Andrew~M Dai. 2018.
\newblock Maskgan: better text generation via filling in the\_.
\newblock \emph{International Conference on Learning Representations}.

\bibitem[{Gong et~al.(2020)Gong, Sun, Feng, Qin, Bi, Liu, and Liu}]{TableGPT}
Heng Gong, Yawei Sun, Xiaocheng Feng, Bing Qin, Wei Bi, Xiaojiang Liu, and Ting
  Liu. 2020.
\newblock Tablegpt: Few-shot table-to-text generation with table structure
  reconstruction and content matching.
\newblock In \emph{International Conference on Computational Linguistics},
  pages 1978–--1988.

\bibitem[{Gu et~al.(2016)Gu, Lu, Li, and Li}]{gu-etal-2016-incorporating}
Jiatao Gu, Zhengdong Lu, Hang Li, and Victor~O.K. Li. 2016.
\newblock \href {https://doi.org/10.18653/v1/P16-1154} {Incorporating copying
  mechanism in sequence-to-sequence learning}.
\newblock In \emph{Proceedings of the 54th Annual Meeting of the Association
  for Computational Linguistics (Volume 1: Long Papers)}, pages 1631--1640,
  Berlin, Germany. Association for Computational Linguistics.

\bibitem[{Guo et~al.(2021)Guo, Tan, Liu, Xing, and Hu}]{Guo2021}
Han Guo, Bowen Tan, Zhengzhong Liu, Eric~P. Xing, and Zhiting Hu. 2021.
\newblock Text generation with efficient (soft) q-learning.
\newblock \emph{arXiv preprint arXiv:2106.07704}.

\bibitem[{He et~al.(2019)He, Zhang, Zhou, and Glass}]{Tianxing2019}
Tianxing He, Jingzhao Zhang, Zhiming Zhou, and James Glass. 2019.
\newblock Exposure bias versus self-recovery: Are distortions really
  incremental for autoregressive text generation?
\newblock \emph{arXiv preprint arXiv:1905.10617}.

\bibitem[{Houlsby et~al.(2019)Houlsby, Giurgiu, Jastrzebski, Morrone,
  De~Laroussilhe, Gesmundo, Attariyan, and Gelly}]{houlsby2019parameter}
Neil Houlsby, Andrei Giurgiu, Stanislaw Jastrzebski, Bruna Morrone, Quentin
  De~Laroussilhe, Andrea Gesmundo, Mona Attariyan, and Sylvain Gelly. 2019.
\newblock Parameter-efficient transfer learning for nlp.
\newblock In \emph{International Conference on Machine Learning}, pages
  2790--2799. PMLR.

\bibitem[{Kale(2020)}]{kale2020text}
Mihir Kale. 2020.
\newblock Text-to-text pre-training for data-to-text tasks.
\newblock \emph{arXiv preprint arXiv:2005.10433}.

\bibitem[{Keneshloo et~al.(2019)Keneshloo, Shi, Ramakrishnan, and
  Reddy}]{keneshloo2019deep}
Yaser Keneshloo, Tian Shi, Naren Ramakrishnan, and Chandan~K Reddy. 2019.
\newblock Deep reinforcement learning for sequence-to-sequence models.
\newblock \emph{IEEE transactions on neural networks and learning systems},
  31(7):2469--2489.

\bibitem[{Khandelwal et~al.(2019)Khandelwal, Clark, Jurafsky, and
  Kaiser}]{khandelwal2019sample}
Urvashi Khandelwal, Kevin Clark, Dan Jurafsky, and Lukasz Kaiser. 2019.
\newblock Sample efficient text summarization using a single pre-trained
  transformer.
\newblock \emph{arXiv preprint arXiv:1905.08836}.

\bibitem[{Konda and Tsitsiklis(2000)}]{konda2000actor}
Vijay~R Konda and John~N Tsitsiklis. 2000.
\newblock Actor-critic algorithms.
\newblock In \emph{Advances in neural information processing systems}, pages
  1008--1014. Citeseer.

\bibitem[{Lazaridou et~al.(2020)Lazaridou, Potapenko, and
  Tieleman}]{lazaridou-etal-2020-multi}
Angeliki Lazaridou, Anna Potapenko, and Olivier Tieleman. 2020.
\newblock \href {https://doi.org/10.18653/v1/2020.acl-main.685} {Multi-agent
  communication meets natural language: Synergies between functional and
  structural language learning}.
\newblock In \emph{Proceedings of the 58th Annual Meeting of the Association
  for Computational Linguistics}, pages 7663--7674, Online. Association for
  Computational Linguistics.

\bibitem[{Lewis et~al.(2020)Lewis, Liu, Goyal, Ghazvininejad, Mohamed, Levy,
  Stoyanov, and Zettlemoyer}]{Mike2020bart}
Mike Lewis, Yinhan Liu, Naman Goyal, Marjan Ghazvininejad, Abdelrahman Mohamed,
  Omer Levy, Ves Stoyanov, and Luke Zettlemoyer. 2020.
\newblock Bart: Denoising sequence-to-sequence pre-training for natural
  language generation, translation, and comprehension.
\newblock In \emph{Proceedings of the 58th Annual Meeting of the Association
  for Computational Linguistics}, pages 7871--7880.

\bibitem[{Li and Liang(2021{\natexlab{a}})}]{Lisa2021}
Xiang~Lisa Li and Percy Liang. 2021{\natexlab{a}}.
\newblock Prefix-tuning: Optimizing continuous prompts for generation.
\newblock \emph{arXiv preprint arXiv:2101.00190}.

\bibitem[{Li and Liang(2021{\natexlab{b}})}]{li-liang-2021-prefix}
Xiang~Lisa Li and Percy Liang. 2021{\natexlab{b}}.
\newblock \href {https://doi.org/10.18653/v1/2021.acl-long.353} {Prefix-tuning:
  Optimizing continuous prompts for generation}.
\newblock In \emph{Proceedings of the 59th Annual Meeting of the Association
  for Computational Linguistics and the 11th International Joint Conference on
  Natural Language Processing (Volume 1: Long Papers)}, pages 4582--4597,
  Online. Association for Computational Linguistics.

\bibitem[{Lin(2004)}]{lin-2004-rouge}
Chin-Yew Lin. 2004.
\newblock \href {https://aclanthology.org/W04-1013} {{ROUGE}: A package for
  automatic evaluation of summaries}.
\newblock In \emph{Text Summarization Branches Out}, pages 74--81, Barcelona,
  Spain. Association for Computational Linguistics.

\bibitem[{Mager et~al.(2020)Mager, Astudillo, Naseem, Sultan, Lee, Florian, and
  Roukos}]{mager2020gpt}
Manuel Mager, Ram{\'o}n~Fernandez Astudillo, Tahira Naseem, Md~Arafat Sultan,
  Young-Suk Lee, Radu Florian, and Salim Roukos. 2020.
\newblock Gpt-too: A language-model-first approach for amr-to-text generation.
\newblock \emph{arXiv preprint arXiv:2005.09123}.

\bibitem[{Nguyen et~al.(2016)Nguyen, Rosenberg, Song, Gao, Tiwary, Majumder,
  and Deng}]{nguyen2016ms}
Tri Nguyen, Mir Rosenberg, Xia Song, Jianfeng Gao, Saurabh Tiwary, Rangan
  Majumder, and Li~Deng. 2016.
\newblock Ms marco: A human generated machine reading comprehension dataset.
\newblock In \emph{CoCo@ NIPS}.

\bibitem[{Paulus et~al.(2017)Paulus, Xiong, and Socher}]{paulus2017deep}
Romain Paulus, Caiming Xiong, and Richard Socher. 2017.
\newblock A deep reinforced model for abstractive summarization.
\newblock \emph{arXiv preprint arXiv:1705.04304}.

\bibitem[{Peng et~al.(2020)Peng, Zhu, Li, Li, Li, Zeng, and Gao}]{peng2020few}
Baolin Peng, Chenguang Zhu, Chunyuan Li, Xiujun Li, Jinchao Li, Michael Zeng,
  and Jianfeng Gao. 2020.
\newblock Few-shot natural language generation for task-oriented dialog.
\newblock \emph{arXiv preprint arXiv:2002.12328}.

\bibitem[{Petroni et~al.(2019)Petroni, Rockt{\"a}schel, Riedel, Lewis, Bakhtin,
  Wu, and Miller}]{petroni-etal-2019-language}
Fabio Petroni, Tim Rockt{\"a}schel, Sebastian Riedel, Patrick Lewis, Anton
  Bakhtin, Yuxiang Wu, and Alexander Miller. 2019.
\newblock \href {https://doi.org/10.18653/v1/D19-1250} {Language models as
  knowledge bases?}
\newblock In \emph{Proceedings of the 2019 Conference on Empirical Methods in
  Natural Language Processing and the 9th International Joint Conference on
  Natural Language Processing (EMNLP-IJCNLP)}, pages 2463--2473, Hong Kong,
  China. Association for Computational Linguistics.

\bibitem[{Radford et~al.(2019)Radford, Wu, Child, Luan, Amodei, and
  Sutskever}]{radford2019language}
Alec Radford, Jeffrey Wu, Rewon Child, David Luan, Dario Amodei, and Ilya
  Sutskever. 2019.
\newblock Language models are unsupervised multitask learners.
\newblock \emph{OpenAI blog}, 1(8):9.

\bibitem[{Ranzato et~al.(2015)Ranzato, Chopra, Auli, and
  Zaremba}]{ranzato2015sequence}
Marc'Aurelio Ranzato, Sumit Chopra, Michael Auli, and Wojciech Zaremba. 2015.
\newblock Sequence level training with recurrent neural networks.
\newblock \emph{arXiv preprint arXiv:1511.06732}.

\bibitem[{Schick and Schütze(2020)}]{Schick2020}
Timo Schick and Hinrich Schütze. 2020.
\newblock Few-shot text generation with pattern-exploiting training.
\newblock \emph{arXiv preprint arXiv:2012.11926}.

\bibitem[{See et~al.(2017)See, Liu, and Manning}]{see2017get}
Abigail See, Peter~J Liu, and Christopher~D Manning. 2017.
\newblock Get to the point: Summarization with pointer-generator networks.
\newblock \emph{arXiv preprint arXiv:1704.04368}.

\bibitem[{Shi et~al.(2021)Shi, Keneshloo, Ramakrishnan, and
  Reddy}]{shi2021neural}
Tian Shi, Yaser Keneshloo, Naren Ramakrishnan, and Chandan~K Reddy. 2021.
\newblock Neural abstractive text summarization with sequence-to-sequence
  models.
\newblock \emph{ACM Transactions on Data Science}, 2(1):1--37.

\bibitem[{Stickland and Murray(2019)}]{Asa2019}
Asa~Cooper Stickland and Iain Murray. 2019.
\newblock Bert and pals: Projected attention layers for efficient adaptation in
  multi-task learning.
\newblock In \emph{International Conference on Machine Learning}, pages
  5986--5995. PMLR.

\bibitem[{Subramanyam~Kalyan et~al.(2021)Subramanyam~Kalyan, Rajasekharan, and
  Sangeetha}]{KalyanAMMUS}
Katikapalli Subramanyam~Kalyan, Ajit Rajasekharan, and Sivanesan Sangeetha.
  2021.
\newblock Ammus : A survey of transformer-based pretrained models in natural
  language processing.
\newblock \emph{arXiv preprint arXiv:2108.05542}.

\bibitem[{Tan et~al.(2018)Tan, Hu, Yang, Salakhutdinov, and Xing}]{Tan2018}
Bowen Tan, Zhiting Hu, Zichao Yang, Ruslan Salakhutdinov, and Eric Xing. 2018.
\newblock Connecting the dots between mle and rl for sequence prediction.
\newblock \emph{arXiv preprint arXiv:1811.09740}.

\bibitem[{Wang et~al.(2020)Wang, Liu, and Song}]{wang2020language}
Chenguang Wang, Xiao Liu, and Dawn Song. 2020.
\newblock Language models are open knowledge graphs.
\newblock \emph{arXiv preprint arXiv:2010.11967}.

\bibitem[{Wu et~al.(2018)Wu, Tian, Qin, Lai, and Liu}]{Lijun2018}
Lijun Wu, Fei Tian, Tao Qin, Jianhuang Lai, and Tie-Yan Liu. 2018.
\newblock A study of reinforcement learning for neural machine translation.
\newblock \emph{arXiv preprint arXiv:1808.08866}.

\bibitem[{Xu et~al.(2021)Xu, Wang, Kim, and Lee}]{AUGNLG}
Xinnuo Xu, Guoyin Wang, Young-Bum Kim, and Sungjin Lee. 2021.
\newblock Augnlg: Few-shot natural language generation using self-trained data
  augmentation.
\newblock \emph{arXiv preprint arXiv:2106.05589}.

\bibitem[{Zahavy et~al.(2018)Zahavy, Haroush, Merlis, Mankowitz, and
  Mannor}]{zahavy2018learn}
Tom Zahavy, Matan Haroush, Nadav Merlis, Daniel~J Mankowitz, and Shie Mannor.
  2018.
\newblock Learn what not to learn: Action elimination with deep reinforcement
  learning.
\newblock \emph{arXiv preprint arXiv:1809.02121}.

\bibitem[{Zeldes et~al.(2020)Zeldes, Padnos, Sharir, and
  Peleg}]{zeldes2020technical}
Yoel Zeldes, Dan Padnos, Or~Sharir, and Barak Peleg. 2020.
\newblock Technical report: Auxiliary tuning and its application to conditional
  text generation.
\newblock \emph{arXiv preprint arXiv:2006.16823}.

\bibitem[{Zheng and Huang(2021)}]{Zheng2021}
Chujie Zheng and Minlie Huang. 2021.
\newblock Exploring prompt-based few-shot learning for grounded dialog
  generation.
\newblock \emph{arXiv preprint arXiv:2109.06513}.

\bibitem[{Ziegler et~al.(2019)Ziegler, Stiennon, Wu, Brown, Radford, Amodei,
  Christiano, and Irving}]{Ziegler2019}
Daniel~M. Ziegler, Nisan Stiennon, Jeffrey Wu, Tom~B. Brown, Alec Radford,
  Dario Amodei, Paul Christiano, and Geoffrey Irving. 2019.
\newblock Fine-tuning language models from human preferences.
\newblock \emph{arXiv preprint arXiv:1909.08593}.

\end{thebibliography}
